\pdfoutput=1
\documentclass[conference]{IEEEtran}
\IEEEoverridecommandlockouts
\usepackage{cite}
\usepackage{amsmath,amssymb,amsfonts}
\usepackage{balance}
\usepackage{graphicx}
\usepackage{textcomp}
\usepackage{xcolor}
\usepackage[normalem]{ulem}
\newcommand{\redstrike}[1]{\textcolor{red}{\sout{#1}}}
\usepackage{subcaption}
\usepackage{hyperref}
\usepackage{booktabs}
\usepackage{xcolor}
\usepackage{algorithm}
\usepackage{algpseudocode} 
\def\BibTeX{{\rm B\kern-.05em{\sc i\kern-.025em b}\kern-.08em
    T\kern-.1667em\lower.7ex\hbox{E}\kern-.125emX}}

\newif\iffinal

\iffinal
    \newcommand{\zhao}[1]{}
\else
    \newcommand\zhao[1]{{\color{cyan}[Zhao: #1]}}
\fi

\begin{document}

\title{SCAPE: Accurate and Efficient LLM Training with Extreme Sparse Communication} 


\author{
Mingkai Zheng, Junlin Chen, Haotian Xie, and Zhao Zhang \\
Rutgers University \\
\texttt{\{mz687, junlin.chen110, haotian.xie, zhao.zhang\}@rutgers.edu}
}

\maketitle

\begin{abstract}
Communication increasingly dominates the cost of Large Language Model (LLM) pre-training, especially under data-parallel and sharded training schemes, where gradient synchronization and parameter reconstruction overhead increase with model size and system scale.
Existing communication-reduction methods either sparsify raw gradients, which can be unstable for modern Adam-style optimizers at high sparsity, or quantize communication, whose savings are fundamentally bounded by bit width and often incur additional runtime overhead. 
We present SCAPE, a communication-efficient distributed optimizer for LLM training that exploits the stability of AdamS's first-moment to enable aggressive sparsification without loss of LLM quality. 
Instead of constructing masks from raw gradients, SCAPE derives them from first-moment-based statistics, partitions mask generation across workers to align with optimizer sharding, and delays mask usage by one step so that mask synchronization can overlap with computation. 
SCAPE also reconstructs the quantities required for second-moment updates from a single synchronized sparse buffer, avoiding an additional collective. 
We implement SCAPE in Megatron-LM and evaluate its convergence by pre-training GPT-345M on OpenWebText and Llama-500M on SlimPajama-6B using 32 NVIDIA GH200 GPUs on TACC Vista. 
In both models, SCAPE preserves training stability, validation loss, and downstream task accuracy under 90\% and 99\% sparsity. 
For Llama-500M, SCAPE reduces end-to-end pre-training wall-clock time by up to 43.3\% while maintaining model quality comparable to dense AdamW and AdamS.
For Llama-1.8B, SCAPE achieves up to 3.26$\times$ speedup per step compared to dense AdamS.
\end{abstract}


\section{Introduction}
\label{sec:intro}

\begin{figure}[t]
    \centering
    \includegraphics[width=0.9\linewidth]{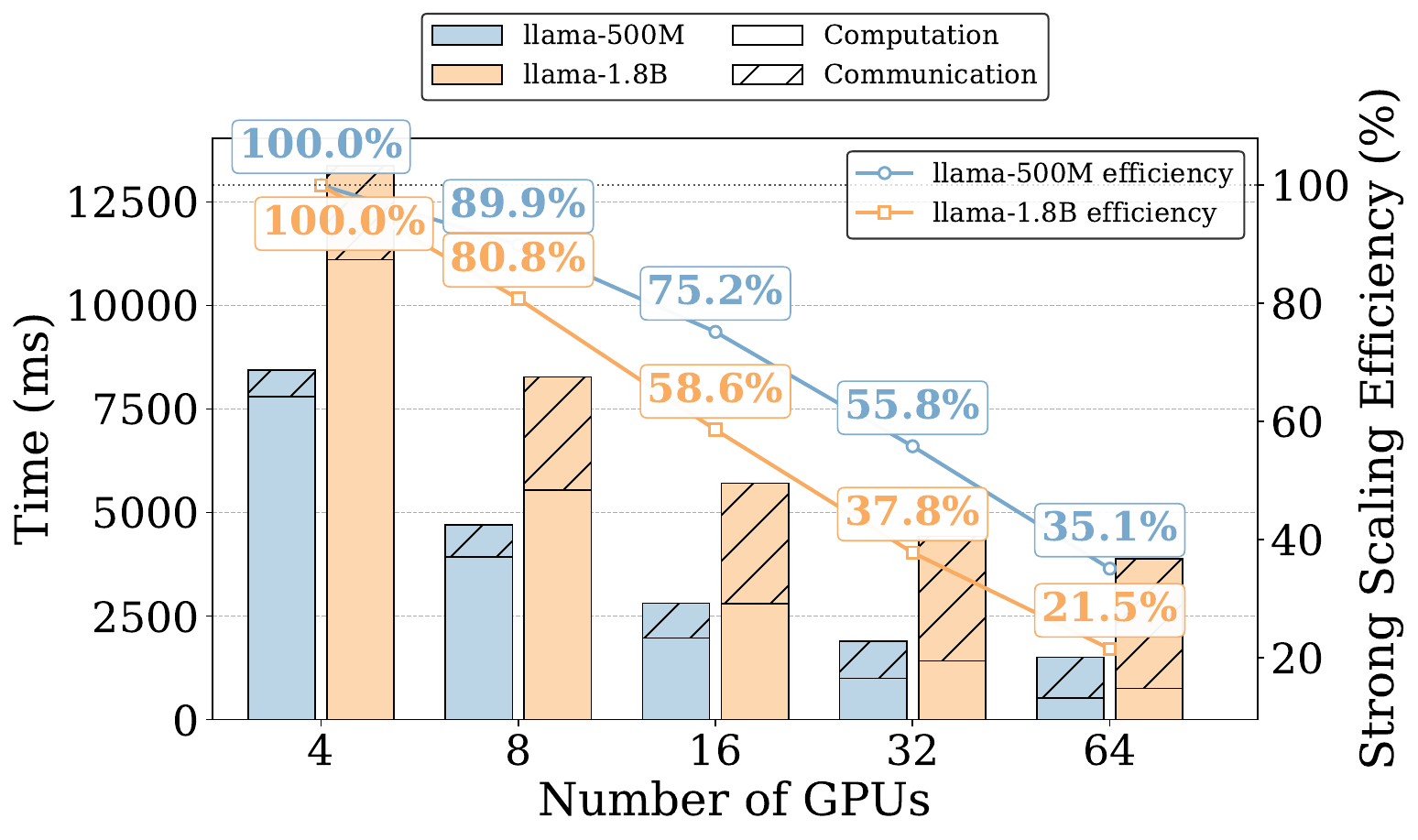}
    \caption{Scaling bottleneck for pre-training Llama-500M (sequence length 4K) and Llama-1.8B (sequence length 2K) using Megatron-LM with distributed optimizer on NVIDIA GH200 GPUs. Llama-1.8B uses a sequence length of 2K due to the limited memory of GH200.}
    \label{fig:scaling-bottleneck}
\end{figure}

Large Language Models (LLMs) with ever-increasing sizes have achieved unparalleled performance across many fields, including math reasoning~\cite{ahn2024largelanguagemodelsmathematical}, code generation~\cite{chen2021evaluating}, and autonomous laboratory~\cite{szymanski2023auto}. 
State-of-the-art LLM training requires a massive amount of graphics processing units (GPUs) to achieve faster training and accommodate enormous model parameters and optimizer states.
The legacy data parallel (DP) strategy replicates the model and optimizer states (e.g., first- and second-momentum in AdamW~\cite{loshchilov2019decoupledweightdecayregularization}) across GPUs and distributes a mini-batch of data among them.
Modern sharded data parallel strategies, such as ZeRO~\cite{rajbhandari2020zeromemoryoptimizationstraining}, FSDP~\cite{zhao2023pytorchfsdpexperiencesscaling}, and Megatron-LM~\cite{megatronlm}, reduce spatial redundancy by partitioning the model and optimizer states across GPUs.
In either case, communication (i.e., \textit{all-reduce} and \textit{all-gather}) is the scaling bottleneck given the dependency on model sizes and the scale, as shown in \autoref{fig:scaling-bottleneck}.



Lowering communication volume is an effective way to reduce communication overhead. 
Researchers have explored several approaches.
DGC~\cite{lin2020deepgradientcompressionreducing}, DeMo~\cite{peng2026demodecoupledmomentumoptimization}, EDGC~\cite{EDGC}, oktopk~\cite{li2022oktopk}, and Radius~\cite{Radius} communicates top-$k$ gradients with error feedback to preserve model performance.
A second line of research exploits quantization (low-bit representation) techniques.
QSDP~\cite{markov2023quantizeddistributedtraininglarge}, ZeRO++~\cite{wang2023zeroextremelyefficientcollective}, and SDP4bit~\cite{jia2024sdp4bit4bitcommunicationquantization} reduce the number of bits for model parameters and gradients in the sharded data parallel strategy.


Existing top-$k$ methods are limited in applicability to modern optimizers or require low sparsity due to constraints on model performance.
Legacy top-$k$ methods, such as DGC~\cite{lin2020deepgradientcompressionreducing} and DeMo~\cite{peng2026demodecoupledmomentumoptimization}, are designed to sparsify the momentum in the SGD~\cite{stich2018sparsifiedsgdmemory} optimizer, which is rarely used in today's LLM training. 
Radius~\cite{Radius} proposes an error-correction mechanism for the AdamW~\cite{loshchilov2019decoupledweightdecayregularization} optimizer, and can achieve 40\% sparsity without degrading downstream task performance. 
The communication volume reduction capability of quantization methods, including QSDP~\cite{markov2023quantizeddistributedtraininglarge}, ZeRO++~\cite{wang2023zeroextremelyefficientcollective}, and SDP4bit~\cite{jia2024sdp4bit4bitcommunicationquantization}, is inherently limited by the number of bits required to represent each gradient.
Furthermore, without NCCL backend support, these quantization-based methods rely on expensive \textit{all-to-all} communication, imposing high memory and communication overhead.



In this work, we ask the following research question: Can we reduce communication volume to the extreme without harming the model performance?
To this end, we propose \textbf{SCAPE}, a communication-efficient distributed optimizer that sparsifies the optimizer states rather than gradients.
SCAPE is built upon AdamS~\cite{zhang-etal-2025-adams} and is inspired with two important insights:
\begin{itemize}
    \item Compared to AdamW, AdamS exhibits substantially smaller residual growth under high sparsity in LLM pretraining, indicating improved robustness to stale error-feedback dynamics.
    \item The top-$k$ structure of the first-moment is temporally stable across two adjacent steps, suggesting that delayed mask reuse is practical.
\end{itemize}
Based on these observations, SCAPE introduces a partitioned mask-refresh mechanism aligned with optimizer sharding.
SCAPE computes masks from momentum-based statistics rather than raw gradients, and delays their use by one step so that mask synchronization can overlap with computation. 
In addition, SCAPE reconstructs the quantities needed for second-moment updates using a single synchronized sparse buffer, avoiding a second \textit{all-reduce}. 
When sharded data parallel is enabled, with the model and optimizer states distributed across all GPUs, SCAPE compresses the volume using sparsity for both \textit{reduce-scatter} for gradient synchronization and \textit{all-gather} for reconstructing the model.


We implement SCAPE using Megatron-LM~\cite{megatronlm} and empirically verify its convergence through pre-training GPT-345M on the OpenWebText~\cite{Gokaslan2019OpenWeb} dataset and Llama-500M on the SlimPajama-6B~\cite{cerebras2023slimpajama} dataset.
We run experiments on the TACC Vista supercomputer with evaluation on an extensive suite of downstream tasks. 
Our experiment results show that SCAPE with 90\% and 99\% sparsity can reduce the wall-clock time for Llama-500M pretraining on 32 GH200 GPUs by 35.6\% and 43.3\%, respectively. 
More importantly, under such high sparsities, SCAPE maintains final training and validation losses close to those of dense AdamS and AdamW, without affecting their performance on downstream benchmarks, such as LAMBADA, SuperGLUE, PIQA, MMLU, and ARC. 

SCAPE is expected to be effective across GPU clusters with various GPU and interconnect configurations, though the improvements it achieves may vary depending on the hardware.
SCAPE leverages NVLink-C2C with 900~GB/s throughput between the CPU and GPU on the GH200 superchips for offloading error feedback buffers.
Buffer offloading may introduce additional overhead on other GPU clusters with PCIe connections.
However, the impact of buffer offloading is minimal compared to the overall improvement with SCAPE.



\section{Background}
In this section, we provide background information on distributed training with Megatron-LM \cite{megatronlm} with distributed optimizers, top-$k$ sparsification for gradient, and the AdamS \cite{zhang-etal-2025-adams} optimizer.

\subsection{Training with Sharded DP Distributed Optimizer}
In the conventional distributed data parallel (DDP) training, each worker (e.g., a GPU) in a data parallel (DP) group duplicates all the model parameters, gradient, and optimizer states, and has a split of a micro-batch of training samples. After all workers finish computing their gradient using the local samples, they synchronize the gradient with a ring \textit{all-reduce} operation before updating the optimizer states and model parameters. Assume there are $N$ workers and the gradient size is $W$ (same as the model size). This communication cost can be decomposed into a \textit{reduce-scatter} following an \textit{all-gather} operation, which is modeled as  
\begin{equation}\label{eq:ring-allreduce}
    \begin{aligned}
    T_\text{ring-$all\text{-}reduce$}&=T_\text{$reduce\text{-}scatter$}+T_\text{$all\text{-}gather$}\\
    &=2(N-1) \left[\alpha + \beta \frac{W}{N}\right],
\end{aligned}
\end{equation}
where $\alpha$ is the fixed latency to initialize a message, and $\beta$ is the inverse of bandwidth. As the number of workers $N$ increases, $2W\beta(N-1)/N$ is dominated by the gradient size $W$ and the inverse of the bandwidth $\beta$.  

\begin{figure}
    \centering
    \includegraphics[width=\linewidth]{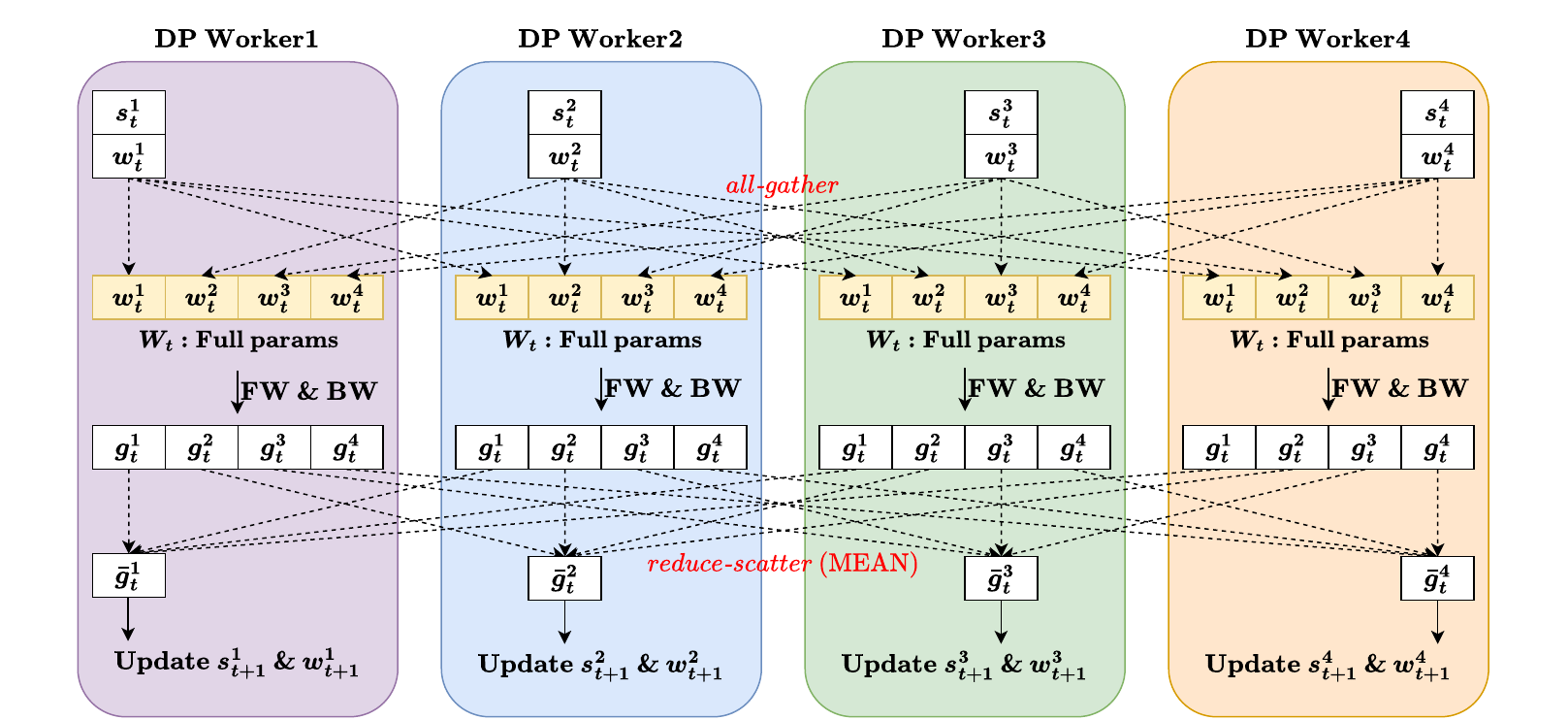}
    \caption{Megatron-LM with sharded data parallel distributed optimizer}
    \label{fig:dist-optm}
\end{figure}

Duplicating the whole model and optimizer states for all workers in a DP group guarantees that the forward and backward computation can finish without communication. However, the optimizer states usually require full precision (FP32) to guarantee the training stability, which becomes a main memory bottleneck in training large language models (LLMs) with increasingly more parameters. Therefore, many frameworks have been proposed to reduce this bottleneck. For instance, Megatron-LM with sharded DP distributed optimizers partitions the optimizer states and the model parameters evenly among all DP workers. Thus, before the forward pass, workers need to perform \textit{all-gather} to collect the model shards from each other. 
Note that the model parameters can be prefetched to overlap the communication of \textit{all-gather} with computation. After the backward pass finishes, each DP worker has the complete gradient. Since the optimizer states have been partitioned evenly, gradients' \textit{all-reduce} synchronization can be replaced by \textit{reduce-scatter} and does not require gathering (i.e., avoid \textit{all-gather}). This process is illustrated in \autoref{fig:dist-optm}.

\subsection{Top-$k$ Gradient Sparsification}
Top-$k$ sparsification is a technique for reducing $T_{\text{ring-}\textit{all-reduce}}$ by reducing $W$ in \autoref{eq:ring-allreduce}, given a fixed network bandwidth $1/\beta$. Specifically, before synchronizing the gradient $g$, we perform a top-$k$ operation to select the $k$ largest-magnitude gradient values, and then synchronize those top-$k$ values using a compressed buffer. This effectively reduces the communication volume from $W$ to $kW$. 

Note that the sparsified $g$ is biased, since the top-$k$ operator drops all but the largest-magnitude entries in gradient $g$ and therefore does not preserve the original gradient in expectation. Consequently, directly applying top-$k$ sparsification at every iteration may introduce accumulated compression error and degrade convergence. To mitigate this issue, prior work usually employs error feedback \cite{stich2018sparsifiedsgdmemory}, which accumulates the unsent residual from the current step and adds it back to the gradient before the next top-$k$ selection. Let $e_t$ denote the residual buffer at step $t$, then the communicated gradient is $\tilde{g}_t=\operatorname{TopK}(g_t + e_t)$, and the residual is updated as $e_{t+1}=g_t + e_t - \tilde{g}_t$. In this way, the information discarded in one iteration is not permanently lost, but is gradually re-injected into later communications, improving optimization stability while retaining the communication savings of sparsification. Nevertheless, this requires a dedicated error feedback buffer with the same size as $g$ and usually in FP32 for preserving the accuracy.

Since each worker's data samples for computing the gradient $g_t$ at step $t$ are distinct, their top-$k$ gradient indices are distinct as well. Thus, directly using \textit{all-reduce} to compute average of the sparsified gradient leads to wrong results. There are two approaches for tackling this issue. The first one is to perform two \textit{all-gather} collectives to collect both the top-$k$ mask and the corresponding values, and the second method is to first use \textit{all-gather} to form a synchronized top-$k$ mask and use \textit{all-reduce} to synchronize the gradient entries selected by the globally synchronized mask. However, the second approach has the fill-in effect. As described in \cite{li2022oktopk}, when the number of workers $N$ is large enough, the synchronized global top-$k$ mask is dense rather than sparse.  

\subsection{AdamS Optimizer}
AdamS \cite{zhang-etal-2025-adams} is a new Adam-like optimizer which, instead of keeping a second-moment $v_t$ as an exponential moving average (EMA) of $g_t^2$, uses the first-moment $m_t$ itself as the preconditioner:
\begin{equation}\label{eq:adams-def}
    \begin{aligned}
        &m_t=\beta_1m_{t-1}+(1-\beta_1)g_t,\\
        &v_t=\beta_2{\color{red}{m_{t-1}^{\odot2}}} + (1-\beta_2)g_t^{\odot2}, \\
        &w_t=w_{t-1}-\eta\left(\frac{m_t}{\sqrt{v_t}+\epsilon} + \lambda w_{t-1}\right),\\
    \end{aligned}
\end{equation}
where $\beta_1$ and $\beta_2$ are the coefficients for computing the running average, $w_t$ is the model parameter at step $t$, $\eta$ is the learning rate, $\lambda$ is the weight decay hyper-parameter, and $\epsilon$ maintains the numerical stability. Note that the update metric in \autoref{eq:adams-def} is simplified for demonstration purposes, and in practice, $m_t$ and $v_t$ should be multiplied by their bias correction terms $1/(1-\beta_1^t)$ and $1/(1-\beta_2^t)$.

Compared to AdamW \cite{loshchilov2019decoupledweightdecayregularization} (i.e., Adam \cite{kingma2017adammethodstochasticoptimization} with weight decay), AdamS has superior performance in terms of training stability, model quality, and memory consumption. First, when using the same set of hyperparameters, AdamS is more robust to noise than AdamW and avoids large spikes in its training loss curve. Since $m_t$ is the EMA of gradient $g_t$, it is a smoother preconditioner than AdamW's $v_t$, which is an EMA of gradient squared $g_t^2$. 
When large gradient occurs in the stable region where gradient norms are usually small, it is more robust to such noise.
Second, AdamS achieves model quality comparable to, and in some cases slightly better than, AdamW. In GPT-2 \cite{radford2019language} pretraining, AdamS closely mirrors AdamW across model scales and even attains slightly lower validation perplexity. In post-training with GRPO \cite{shao2024deepseekmathpushinglimitsmathematical} on the Countdown task, AdamS yields score curves that closely align with those of AdamW and occasionally surpass its validation performance. Finally, AdamS eliminates the need to store second moment, reducing optimizer state memory by 50\% and leading to practical system benefits such as lower memory consumption in distributed training (e.g., FSDP \cite{zhao2023pytorchfsdpexperiencesscaling}) and higher throughput in memory-bound large-scale pretraining.

\section{Observations}\label{sec:ob}

In this section, we present our observations when pre-training the GPT-345M model. Based on these observations, we provide insights that form the foundations for our system design in \autoref{sec:design}.

\subsection{Gradient Distribution}
\begin{figure}[h]
  \centering
  \begin{subfigure}[b]{0.5\columnwidth}
    \centering
    \includegraphics[width=\linewidth]{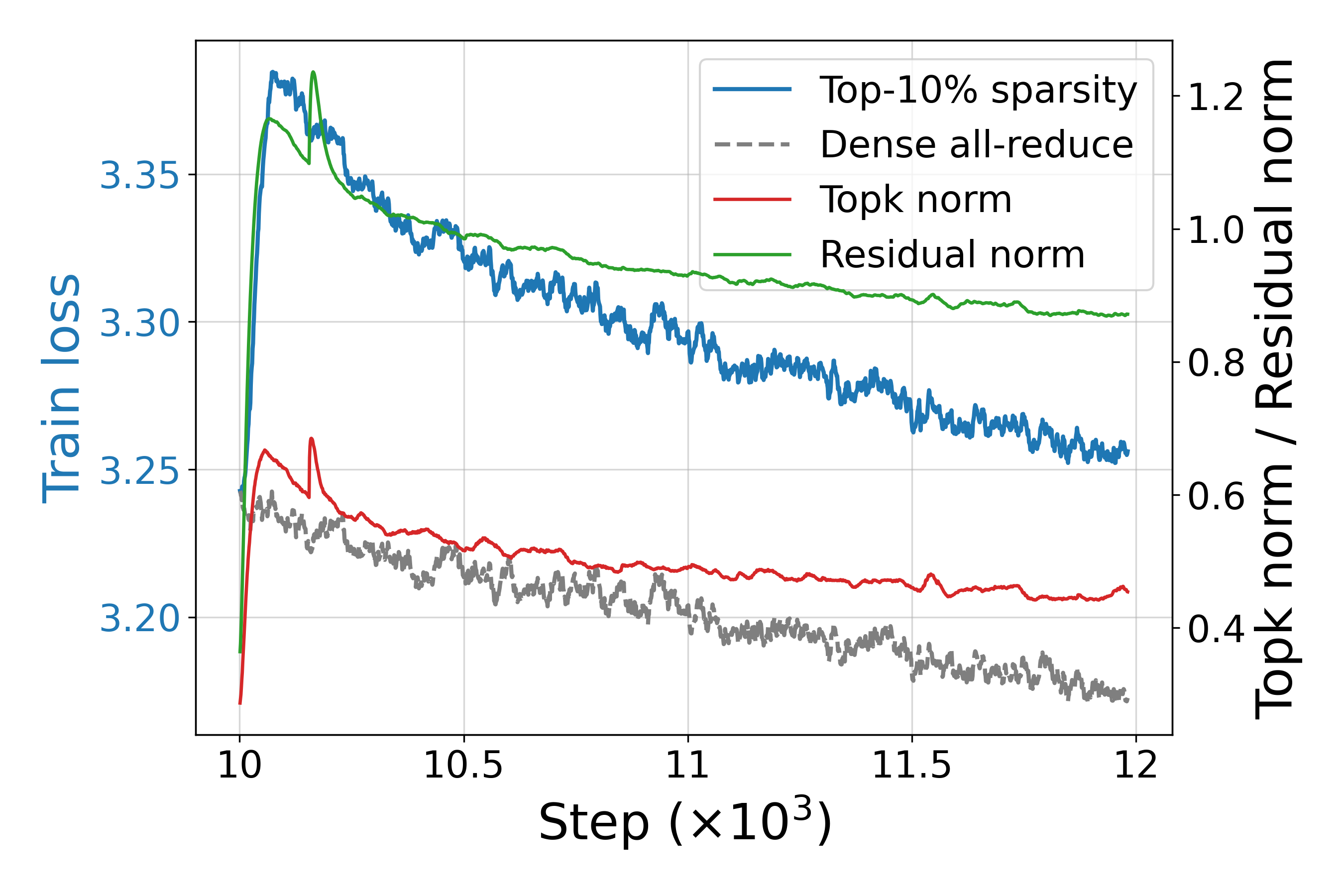}
    \caption{AdamW}
    \label{fig:insight-loss-topk-residual-norm-adamw}
  \end{subfigure}\hfill
  \begin{subfigure}[b]{0.5\columnwidth}
    \centering
    \includegraphics[width=\linewidth]{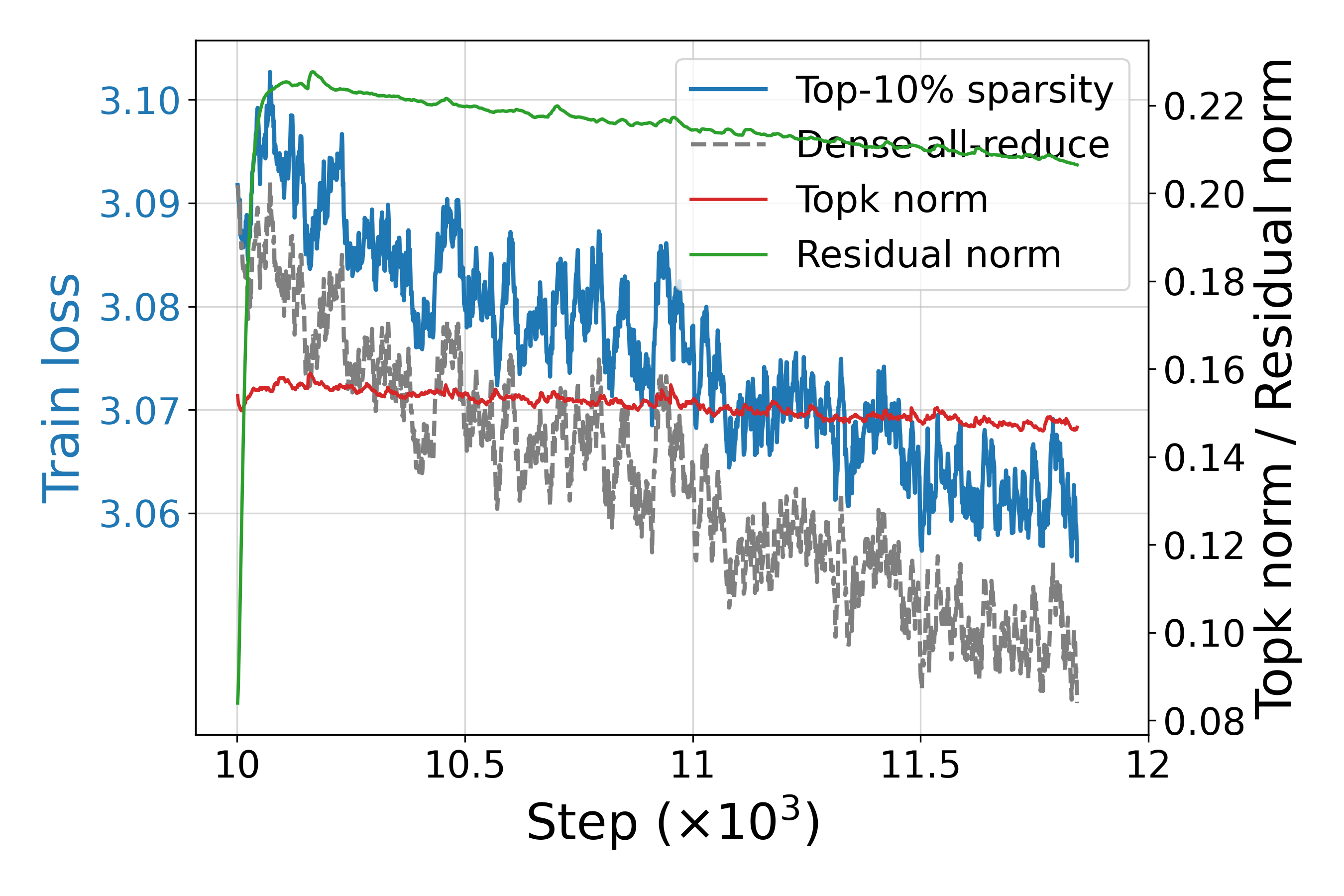}
    \caption{AdamS}
    \label{fig:insight-loss-topk-residual-norm-adams}
  \end{subfigure}
  \caption{Comparison between AdamW and AdamS after switching from dense \textit{all-reduce} to top-10\% sparsified gradient at step 10,000 of pre-training GPT-345M}
  \label{fig:insight-loss-topk-residual-norm}
\end{figure}

We first pre-train GPT-345M with AdamW optimizer on the OpenWebText \cite{Gokaslan2019OpenWeb} dataset with top-10\% sparsity on the gradient, meaning that we only synchronize the gradient with the top-10\% largest magnitudes. In this setup, we switch from a dense \textit{all-reduce} with no gradient compression (baseline method) to top-$10\%$ sparsification at step 10,000. Moreover, the top-$k$ sparsification is performed in a per-layer style. We can see from \autoref{fig:insight-loss-topk-residual-norm-adamw} that the difference between top-$10\%$ sparsity and the baseline method rapidly increases and then slowly decreases. We then try AdamS using the same setup, and the result is presented in \autoref{fig:insight-loss-topk-residual-norm-adams}. We notice that, using AdamS, the gap between the baseline method and the sparsified gradient is much smaller.

To analyze these phenomena, we plot the norms of the top-$k$ sparsified gradient and the residual buffer, which stores the non-top-$k$ gradient accumulated during training. From the curves of top-$k$ gradient norm and the residual norm plotted in \autoref{fig:insight-loss-topk-residual-norm-adamw} and \autoref{fig:insight-loss-topk-residual-norm-adams}, we see that AdamW's residual norm is much higher than AdamS's. With a high sparsity rate, such as 10\% in this experiment, the staleness effect of error feedback can significantly influence AdamW's training process. The reason is that the chance of being selected as top-$k$ is so low that many gradient values with very large magnitudes (but still smaller than the top-10\% threshold) have to wait until they have accumulated for several steps. When such large stale gradient values are added back to the optimizer states, including the first- and second-moment, they cause the optimizer states to drift from the baseline's direction and thus lead to a gradually larger gap between the baseline and top-$k$ method. This aligns well with the findings in~\cite{molybog2023theoryadaminstabilitylargescale} and~\cite{bai2025adaptivepreconditionerstriggerloss} (i.e., the loss spike is triggered by suddenly having a large gradient when the training regime enters a region where the square of the gradient and the second-moment are both small).

\begin{figure}[h]
  \centering
  \begin{subfigure}[b]{0.49\columnwidth}
    \centering
    \includegraphics[width=\linewidth]{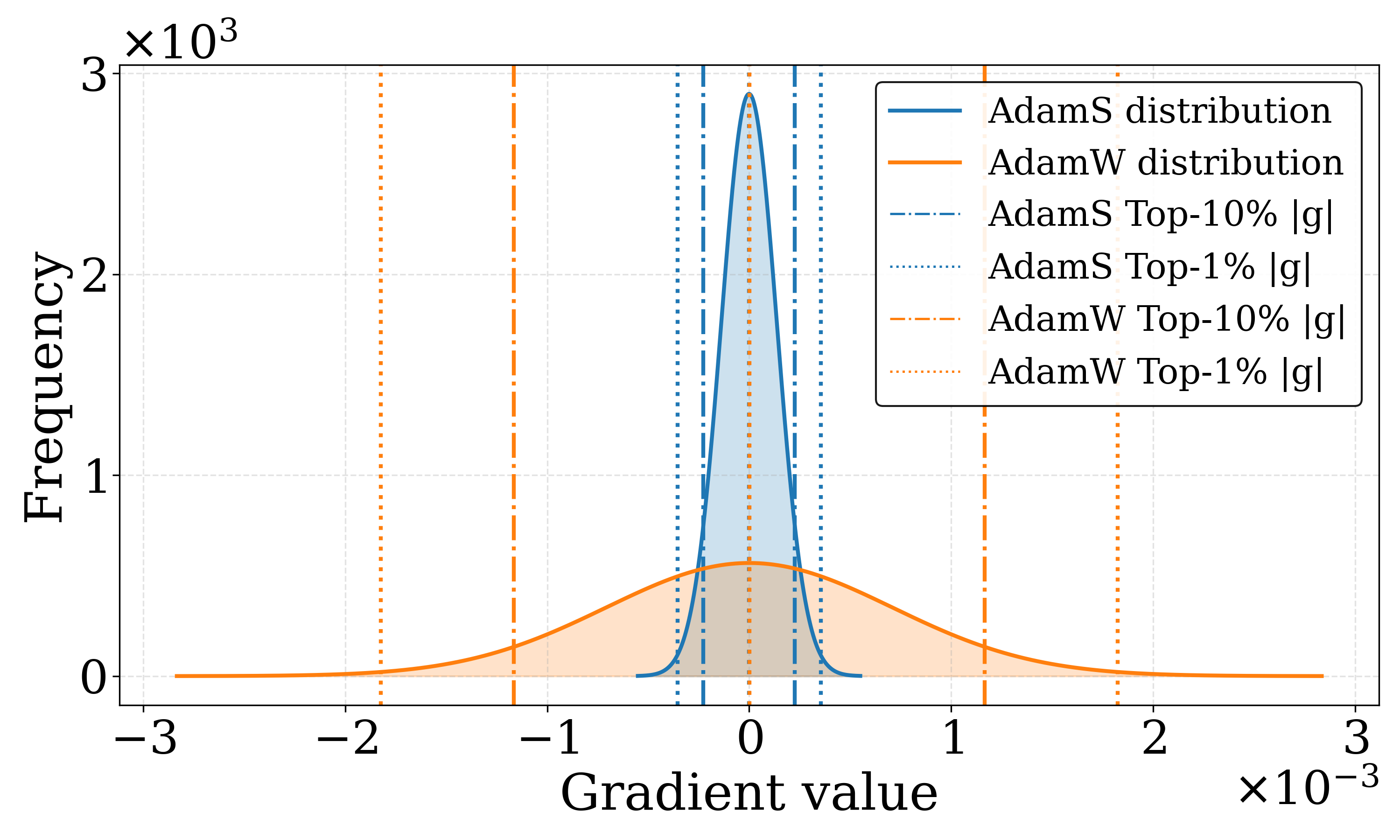}
    \caption{Decoder 0 - Self attention dense at step 1,000}
    \label{fig:grad-distribution-step-1k-layer006}
  \end{subfigure}\hfill
  \begin{subfigure}[b]{0.49\columnwidth}
    \centering
    \includegraphics[width=\linewidth]{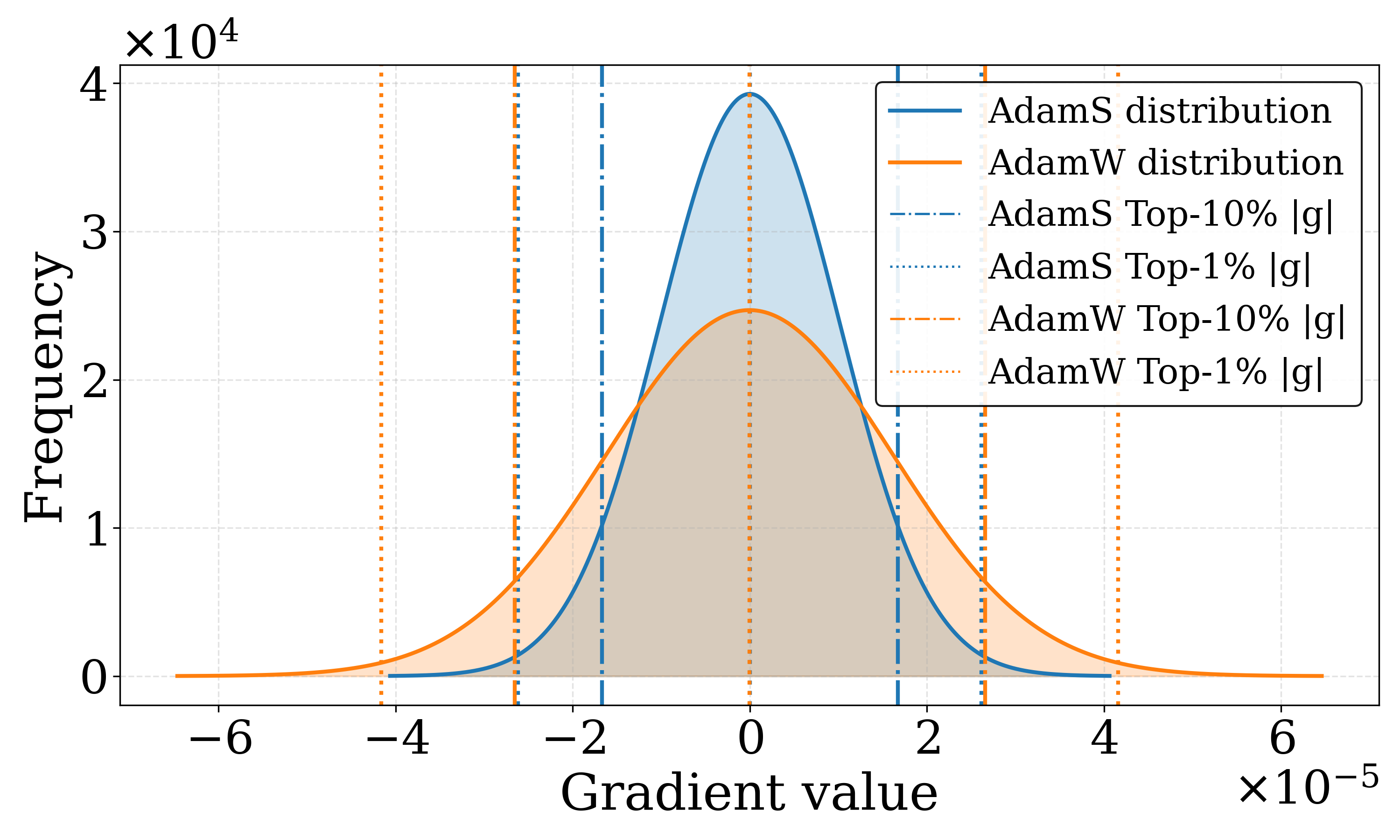}
    \caption{Decoder 10 - MLP $d_\text{model}$ to $4d_\text{model}$ at step 1,000}
  \end{subfigure}
  \begin{subfigure}[b]{0.49\columnwidth}
    \centering
    \includegraphics[width=\linewidth]{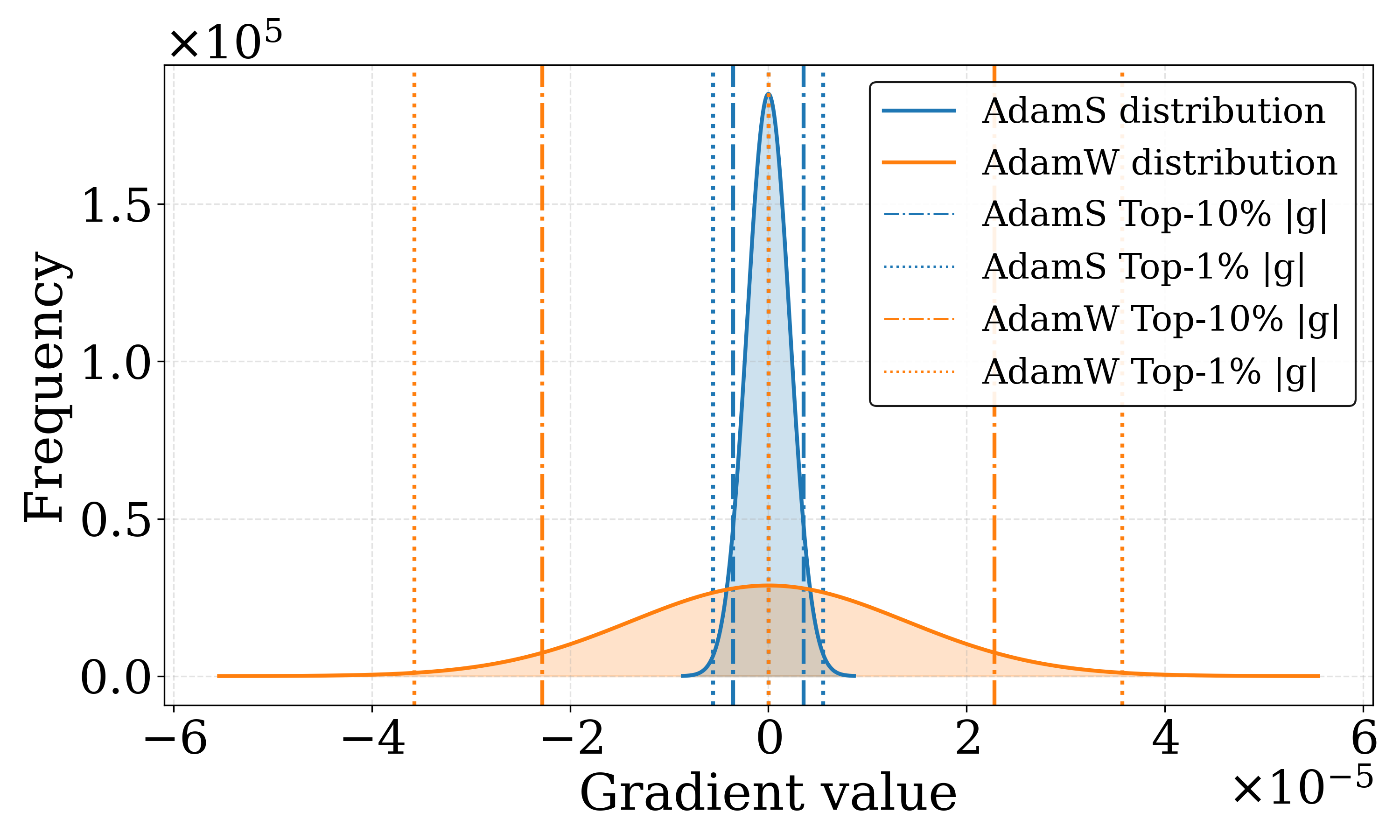}
    \caption{Decoder 0 - Self attention dense at step 50,000}
  \end{subfigure}\hfill
  \begin{subfigure}[b]{0.49\columnwidth}
    \centering
    \includegraphics[width=\linewidth]{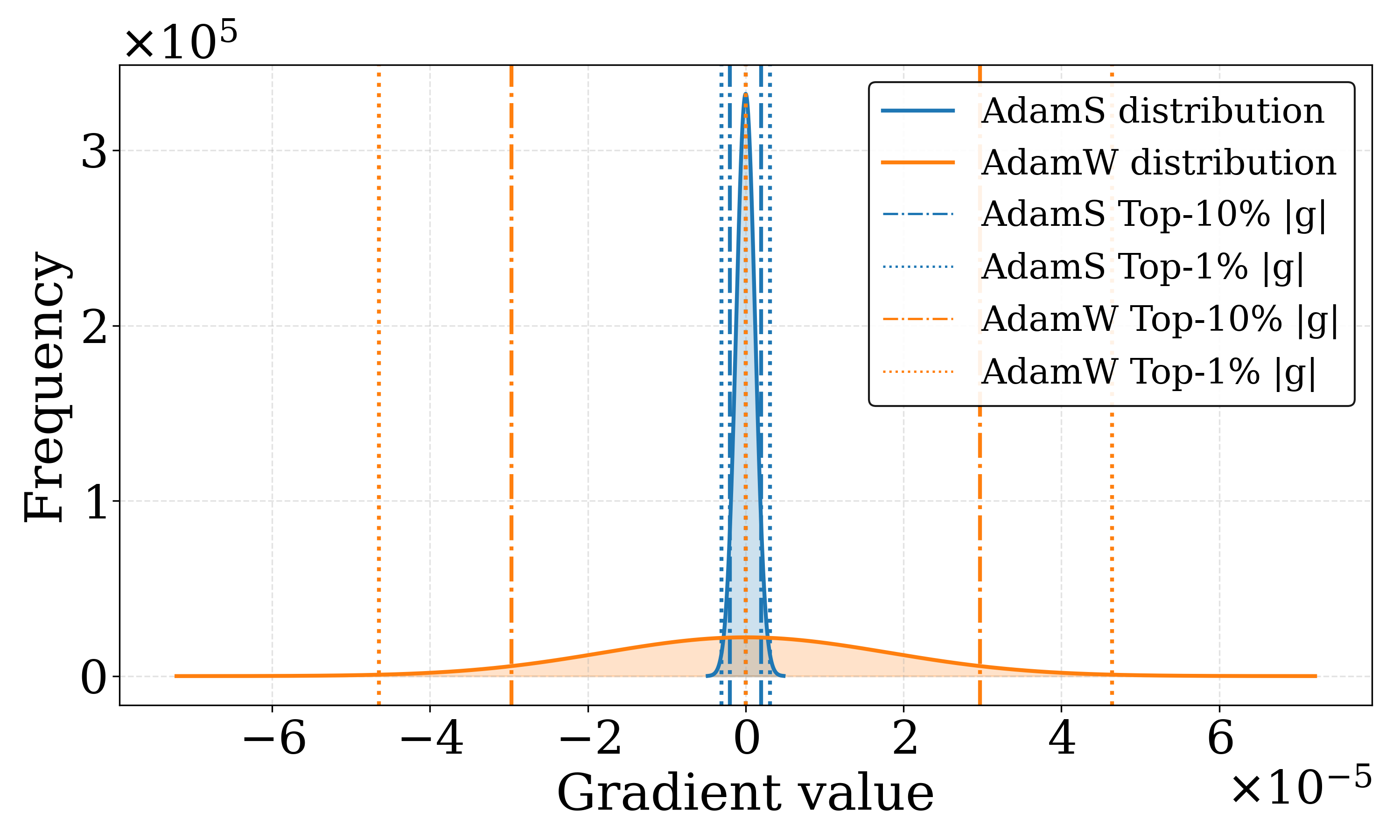}
    \caption{Decoder 2 - QKV projection weight at step 50,000}
  \end{subfigure}
  \caption{Gradient distribution of different layers in GPT-345M at different steps}
  \label{fig:gpt-grad-distribution}
\end{figure}

\begin{figure}[h]
  \centering
  \begin{subfigure}[b]{0.49\columnwidth}
    \centering
    \includegraphics[width=\linewidth]{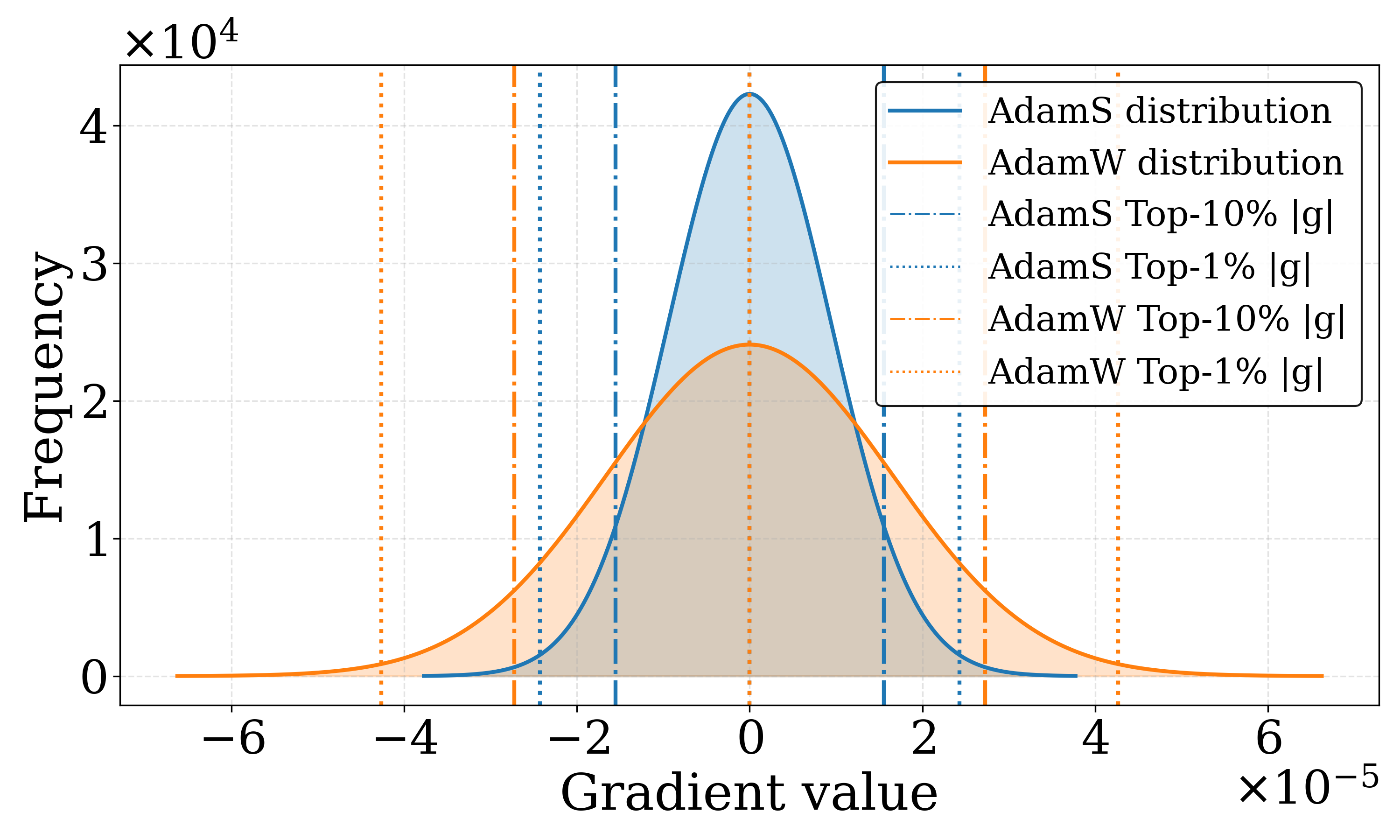}
    \caption{Decoder 0 - gate and up project in SwiGLU at step 1,000}
    \label{fig:gpt-grad-distribution-step-1k-layer006}
  \end{subfigure}\hfill
  \begin{subfigure}[b]{0.49\columnwidth}
    \centering
    \includegraphics[width=\linewidth]{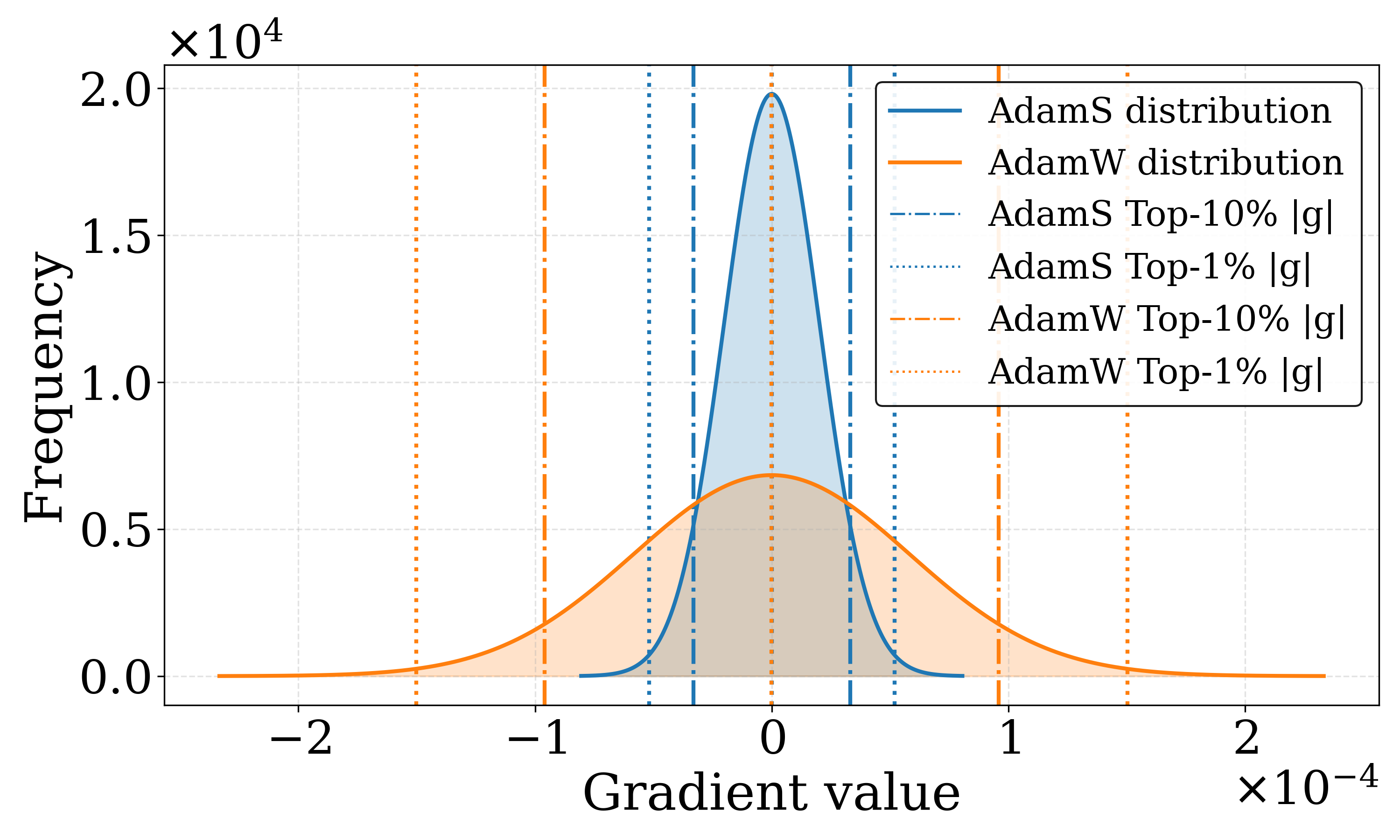}
    \caption{Decoder 8 - Self attention dense at step 1,000}
  \end{subfigure}
  \begin{subfigure}[b]{0.49\columnwidth}
    \centering
    \includegraphics[width=\linewidth]{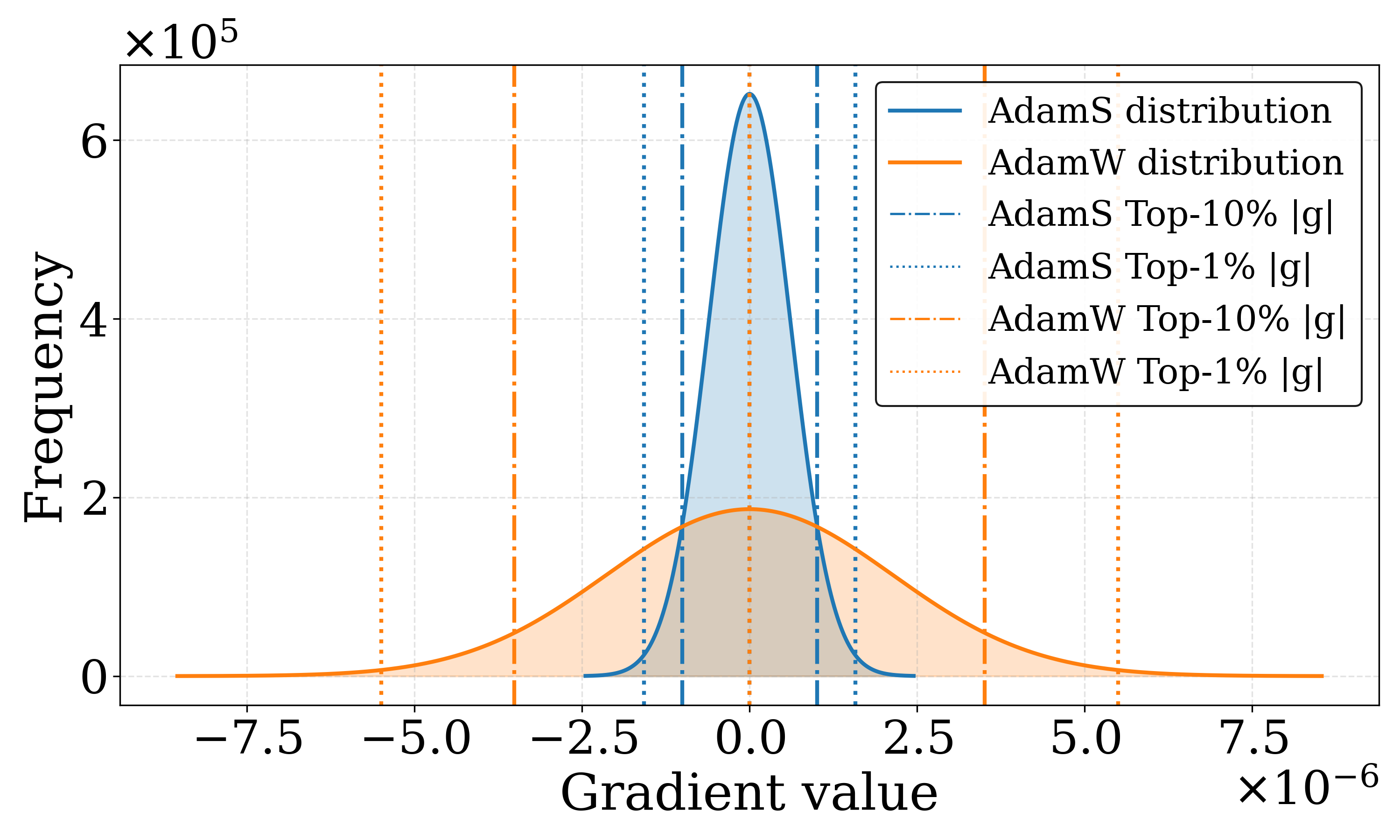}
    \caption{Decoder 0 - gate and up project in SwiGLU at step 50,000}
  \end{subfigure}\hfill
  \begin{subfigure}[b]{0.49\columnwidth}
    \centering
    \includegraphics[width=\linewidth]{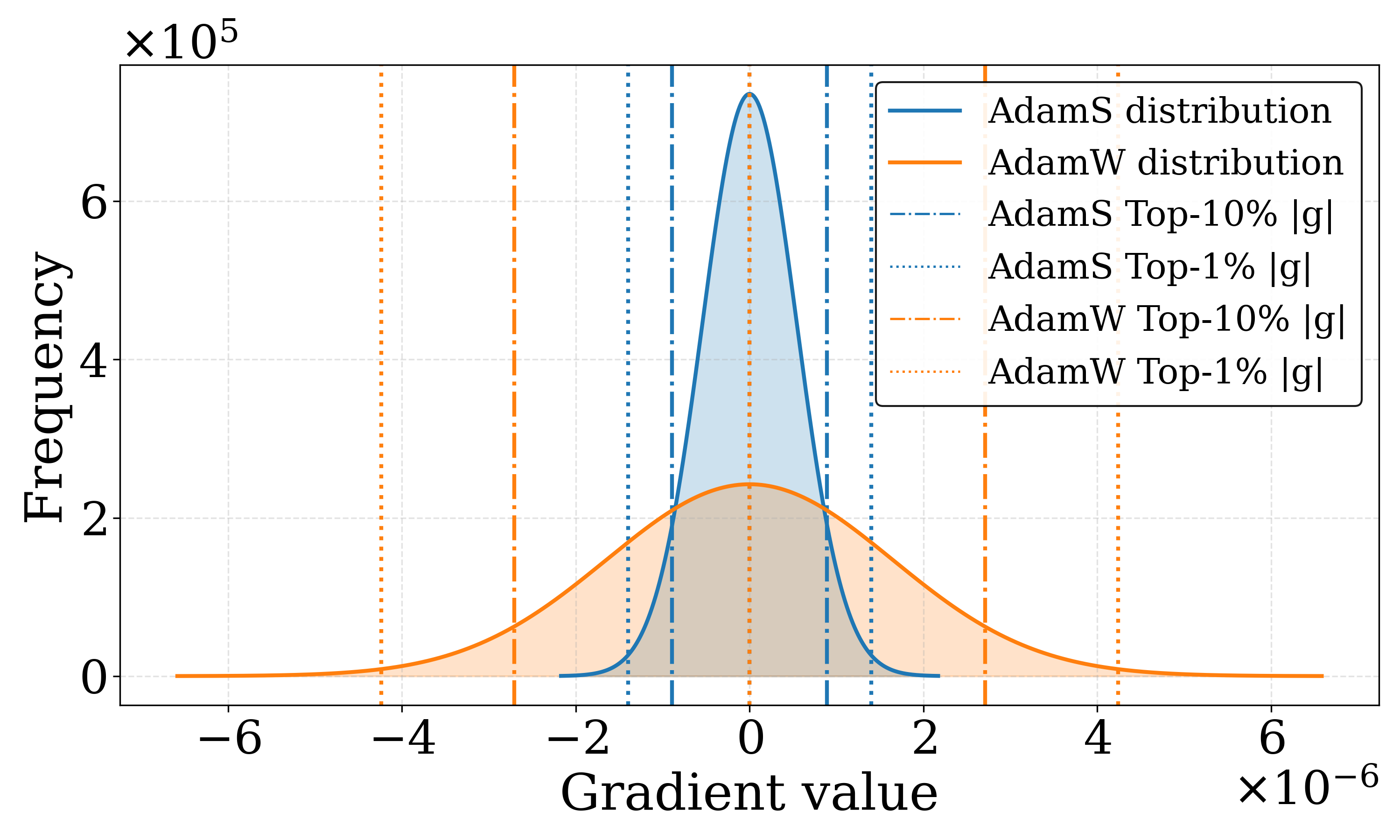}
    \caption{Word embedding at step 50,000}
  \end{subfigure}
  \caption{Gradient distribution of different layers in Llama-500M at different steps}
  \label{fig:llama-grad-distribution}
\end{figure}

However, for AdamS, the norm of the top-$k$ gradient and the norm of the residual values are small, so when adding them back, it does not lead to a significant shift in the optimization direction. We plot the distribution of AdamW's and AdamS's gradient for different layers at two different steps in \autoref{fig:gpt-grad-distribution} for GPT-345M model. We also provide the gradient distribution for Llama-500M in \autoref{fig:llama-grad-distribution}. From these results, we can see that AdamS's gradient is more centralized around 0, while AdamW's gradient distribution spreads out more evenly and is flatter. This suggests that compared to AdamW, AdamS is more suitable for adapting to top-$k$ gradient sparsification.

\subsection{Temporal Stability of first-moment-Derived Top-$k$ Masks}
Prior work has shown that sparse support patterns can exhibit temporal correlation across adjacent training steps. For example, the temporal stability of AdamW's top-1\% gradient has been studied in \cite{Radius}, and \cite{EDGC} reports strong gradient correlation during the early stage of LLM pre-training.

\begin{figure}[hptb]
    \centering
    \begin{subfigure}[b]{0.49\columnwidth}
        \centering
        \includegraphics[width=\linewidth]{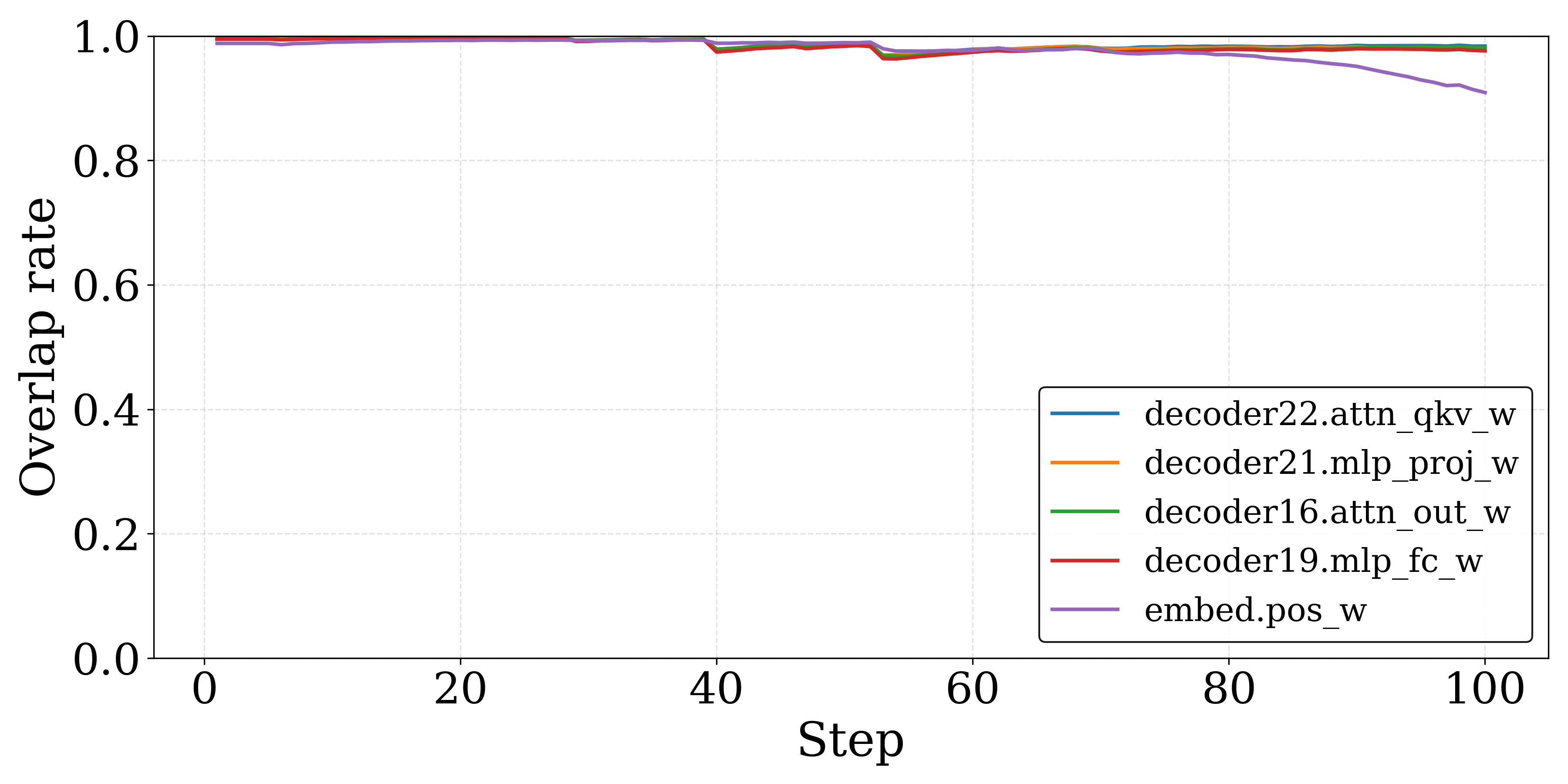}
        \caption{GPT-345M from step 1}
        \label{fig:overlap-rate-gpt-step1}
    \end{subfigure}\hfill
    \begin{subfigure}[b]{0.49\columnwidth}
        \centering
        \includegraphics[width=\linewidth]{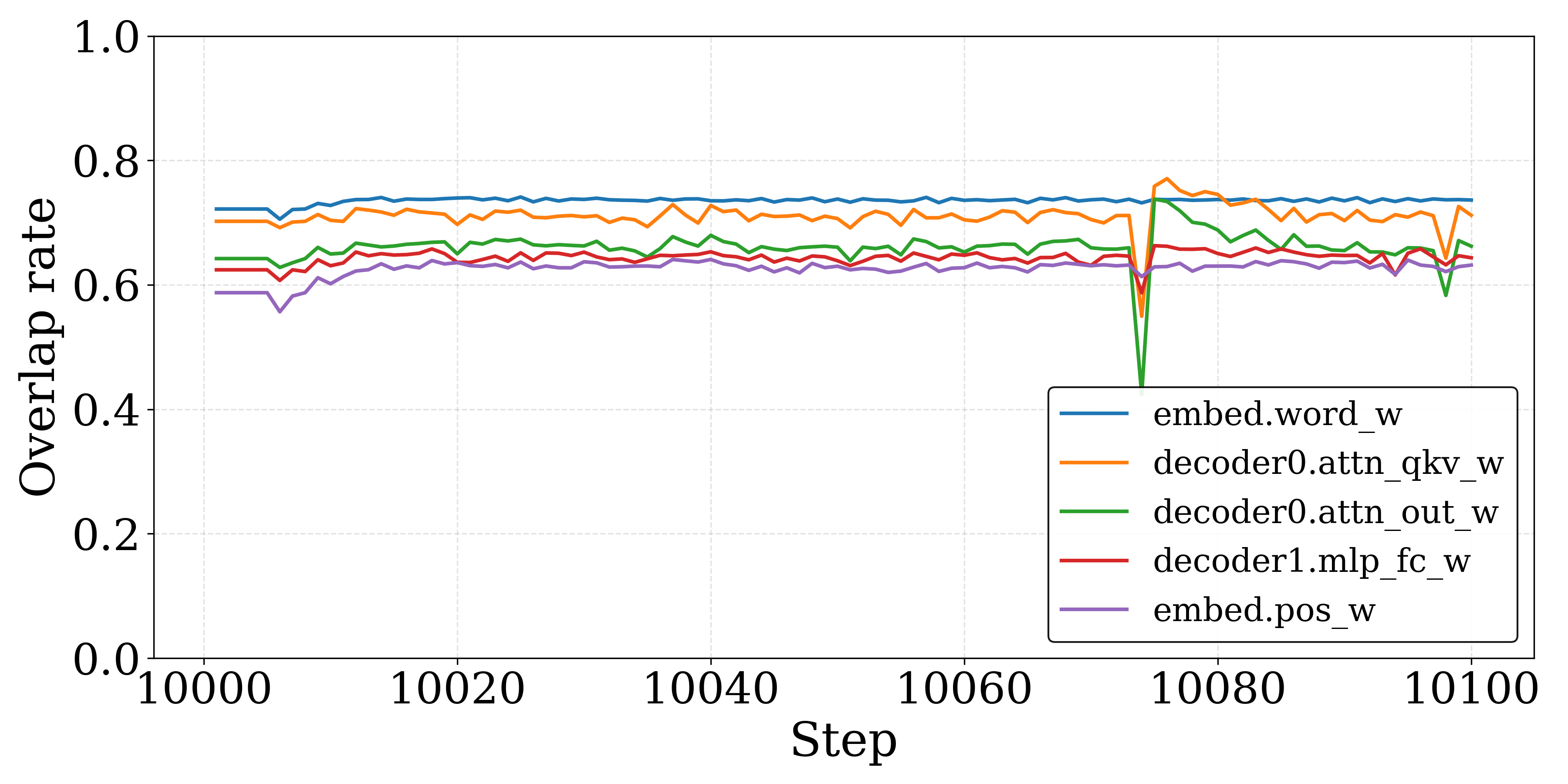}
        \caption{GPT-345M from step 10,000}
        \label{fig:overlap-rate-gpt-step10000}
    \end{subfigure}\hfill
    \begin{subfigure}[b]{0.49\columnwidth}
        \centering
        \includegraphics[width=\linewidth]{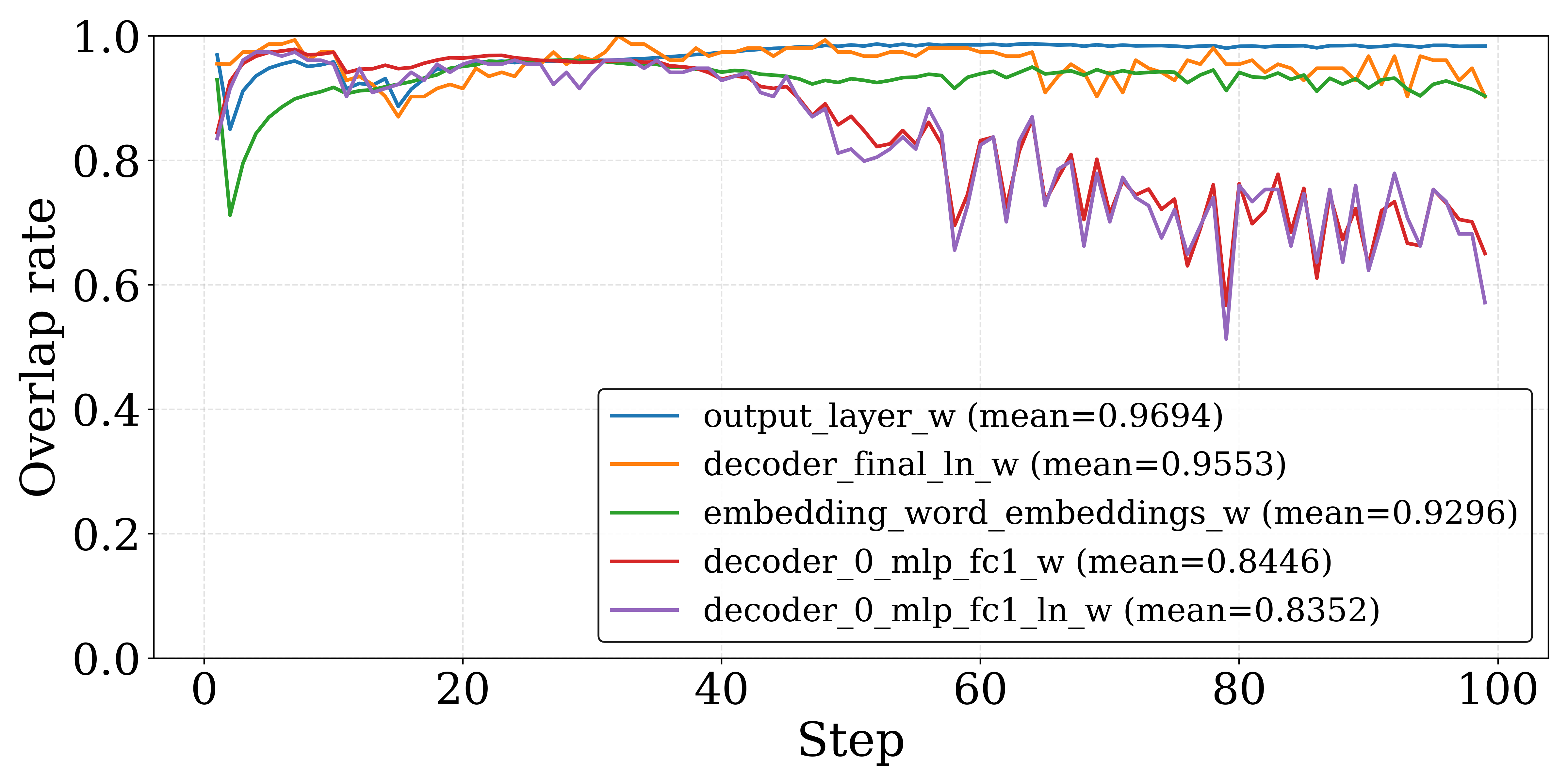}
        \caption{Llama-500M from step 1}
        \label{fig:overlap-rate-llama-step1}
    \end{subfigure}\hfill
    \begin{subfigure}[b]{0.49\columnwidth}
        \centering
        \includegraphics[width=\linewidth]{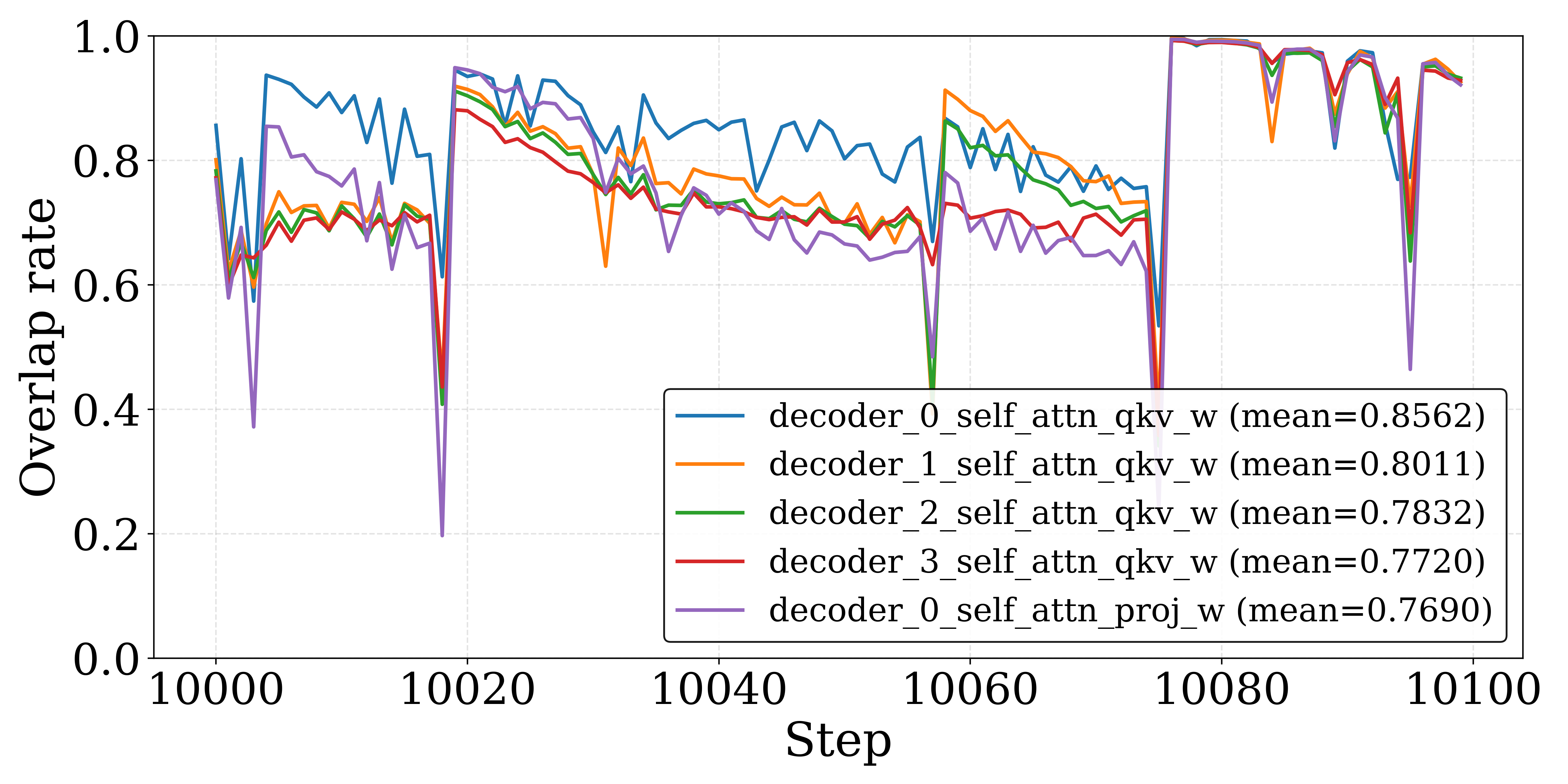}
        \caption{Llama-500M from step 10,000}
        \label{fig:overlap-rate-llama-step10000}
    \end{subfigure}
    \caption{The overlap rate of AdamS's first-moment-derived top-10\% mask between two adjacent steps for different stages of pre-training GPT-345M and Llama-500M} 
    \label{fig:overlap-rate}
\end{figure}

Nevertheless, in our setting, we observe that this property also holds when top-$k$ masks are constructed from AdamS's first-moment rather than raw gradients, for both GPT-345M and Llama-500M. In \autoref{fig:overlap-rate}, we show the overlap between adjacent-step top-10\% masks selected according to the magnitude of AdamS's first-moment across different layers and training stages. As shown in \autoref{fig:overlap-rate-gpt-step1}, many layers exhibit high mask overlap at the beginning of training. Although the overlap gradually decreases as training proceeds, it remains substantial throughout pre-training, indicating that the top-$k$ mask from step $t$ is a useful heuristic for step $t+1$. For Llama-500M, the overlap curves in \autoref{fig:overlap-rate-llama-step1} and \autoref{fig:overlap-rate-llama-step10000} are noisier, but the average overlap for each layer remains high across training stages.



\section{System Design}\label{sec:design}

Motivated by the observations in \autoref{sec:ob}, we design our system in \autoref{algo:main}, which is mainly composed of three important components: 1) new mask synchronization, 2) top-$k$ mask computation, and 3) optimizer state update.

\begin{figure*}[t]
  \centering
  \includegraphics[width=0.9\textwidth]{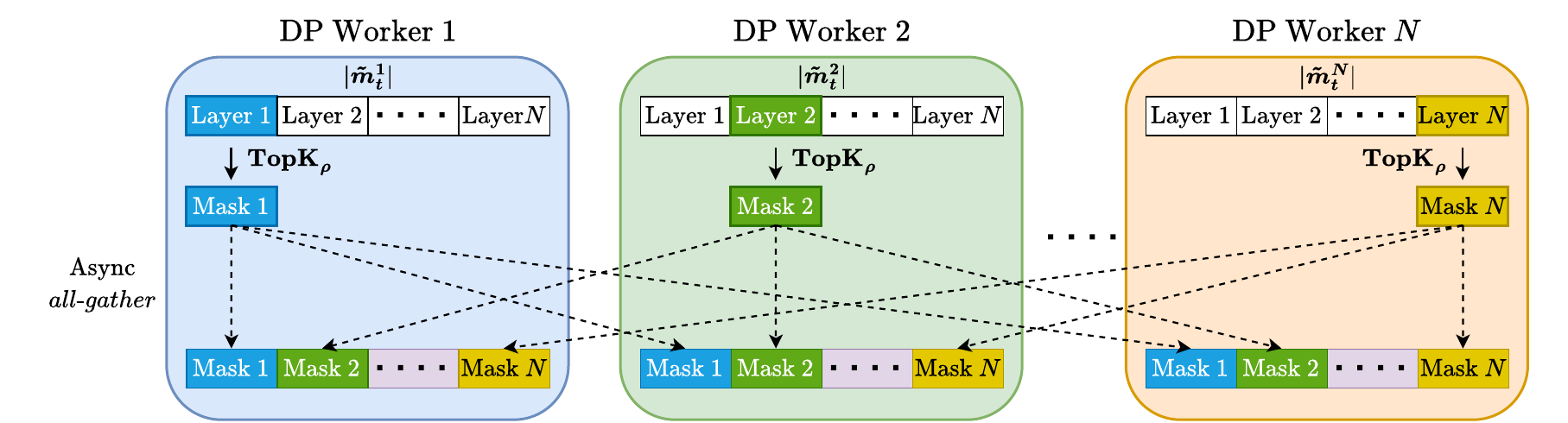}
  \caption{Communication in refreshing top-$k$ mask. Each work computes a sharded top-$k$ mask and then uses \textit{all-gather} to construct the full mask. Since the usage of top-$k$ mask is delayed by one step, asynchronously \textit{all-gather} can be hidden by expensive backward computation.}
  \label{fig:refresh-mask}
\end{figure*}

\begin{algorithm}
  \caption{SCAPE}
  \label{algo:main}
  \begin{algorithmic}[1]
    \Require model parameters $\theta$, dataset $D$, loss function $f$
    \Require step $t$, worker rank $n$, total number of workers $N$
    \Require density $\rho$, weight decay $\lambda$, learning rate $\eta$
    \Require number of layers in model $L$
    \State $M_0\gets \mathbf{1}$ \Comment{Initialize sparse mask}
    \State $e_0^n\gets\mathbf{0}$ \Comment{Initialize EF buffer}
    \For{$t=1,2,\dots$}
    \State $b_t^n\gets0$ \Comment{Re-initialize payload buffer}
    \State $g_t^n\gets \nabla f(D_t^n,\theta_t)$
    \State $\tilde{m}_{t}^n\gets \beta_1m_{t-1} + (1-\beta_1)g_t^n+e_{t-1}^n$
    \State $e_t^n[\neg M_{t-1}] \gets \tilde{m_{t}}[\neg M_{t-1}]$ 
    \State $e_t^n[M_{t-1}] \gets 0$
    \State $b_t^n[M_{t-1}] \gets \tilde{m}_{t}^n[M_{t-1}]$
    \State All-reduce $\bar{b}_t^n$: $\bar{b}_t \gets \frac{1}{N} \sum_{n=1}^N b_t^n$
    \State  $M_t\gets \text{refresh\_mask}(\rho, |\tilde{m}_{t}^n|)$ \Comment{Check \autoref{fig:refresh-mask}}
    \State $m_t\gets \bar{b}_t$ \Comment{First-moment's nontopk all 0}
    \State $\bar{b}_t[M_{t-1}]\gets (\bar{b}_t[M_{t-1}]-\beta_1m_{t-1}[M_{t-1}])/(1-\beta_1)$ 
    \State clip\_grad($\bar{b}_t$)
    \State $v_t\gets \beta_2m_{t-1}^{\odot2} + (1-\beta_2)\bar{b}_t^{\odot2}$
    \State $u_t\gets m_t/(\sqrt{v_t} + \epsilon)$ \Comment{Compute update metrics}
    \State $\theta_{t+1}\gets\theta_{t}-\eta(u_t + \lambda\theta_t)$
    \EndFor
  \end{algorithmic}
\end{algorithm}

\subsection{Refreshed Mask Synchronization}
We propose a new mask refresh strategy illustrated in \autoref{fig:refresh-mask}. This method has two important differences from the existing work. First, instead of having all workers compute their own top-$k$ masks for all layers, each worker only computes a portion. This modification aligns with the design of Megatron-LM's distributed optimizer, where the optimizer states are partitioned evenly onto all workers, and at each step, each worker updates its own portion only. Moreover, when the distributed optimizer is not used, this design reduces each worker's workload of performing the top-$k$ operations by $N$ (i.e., the total number of workers in a DP group). This design is also immune to the fill-in effect mentioned in \cite{li2022oktopk}: as the number of workers grows, the gradient `sparsified' by the collected top-$k$ indices from all workers is nearly dense. Since each worker in our design treats the top-$k$ masks for layers computed by its peers as all zeros, this avoids the fill-in effect. Second, the top-$k$ masks computed from step $t$ are later used at step $t+1$. This delayed use of the top-$k$ masks provides an opportunity to use asynchronous communication to hide the synchronization communication in heavy computations, such as the forward- and backward-pass.

\subsection{Top-$k$ Mask Computation}
The top-$k$ mask can be computed from the gradient and moment. 
SCAPE selects the first-moment to construct the top-$k$ mask. 
The reason is that, as the gradient contains noise inherited from the data, the top-$k$ values generated from the gradient of the current step can have a low overlap rate with the real top-$k$ gradient's indices in the next step. Therefore, we use the first-moment to generate the top-$k$ mask, as it maintains a running average of the gradient, which is more stable and has much lower noise. Specifically, we use $|\tilde{m}_{t}^n|=|\beta_1m_{t-1}+(1-\beta_1)g_t^n+e_{t-1}^n|$ to construct the top-$k$ masks. Because $g_t^n$ and $e_{t-1}^n$ are not synchronized and contain local information, each worker's top-$k$ mask guarantees that its local largest values in $|\tilde{m}_{t}^n|$ are immediately selected and then used to update the optimizer states and model parameters.

\subsection{Optimizer State Update}
In \autoref{algo:main}, SCAPE transmits a sparsified buffer $b_t^n$, which has the values in the temporal, semi-updated first-moment buffer $\tilde{m}_t^n$ on positions selected by the top-$k$ mask $M_{t-1}$. For DGC~\cite{lin2020deepgradientcompressionreducing} and DeMo~\cite{peng2026demodecoupledmomentumoptimization}, transmitting the sparsified first-moment is sufficient to update the optimizer state, because they use SGD optimizer~\cite{robbins1951stochastic}. However, from \autoref{eq:adams-def}, to update the second-moment, we need both the synchronized first-moment and the gradient. A naive implementation would perform two \textit{all-reduce} operations to synchronize both. Nevertheless, we noticed that one \textit{all-reduce} suffices. Given that $m_t$ is guaranteed to be synchronized, we can compute the globally averaged gradient and residual as 
\begin{equation}
    \bar{g}_t+\bar{e}_{t-1}=(\bar{b}_t-\beta_1m_{t-1})/(1-\beta_1).
\end{equation}
Then, we use $\bar{g}_t+\bar{e}_{t-1}$ as the synchronized top-$k$ to update the second-moment $v_t$ and the model parameters $w_t$.

\subsection{Optimizations for Distributed Optimizers}
We can see from \autoref{algo:main} that since $\bar{b}_t$ only has nonzero values for indices selected by top-$k$ mask $M_{t-1}$, we have
\begin{equation}
    m_t[\neg M_{t-1}]=\bar{b}_t[\neg M_{t-1}]=0.
\end{equation}
Therefore, when computing the update metrics, we have 
\begin{equation}
    u_t[\neg M_{t-1}]=m_t[\neg M_{t-1}] / (\sqrt{v_{t}[\neg M_{t-1}]} + \epsilon)=0.
\end{equation}
Hence, $\theta_{t+1}[\neg M_{t-1}]$ is updated as
\begin{equation}
    \begin{aligned}
        \theta_{t+1}[\neg M_{t-1}] 
        &= \theta_t[\neg M_{t-1}]-\eta(u_t[\neg M_{t-1}] + \lambda\theta_t[\neg M_{t-1}])\\
        &= \theta_t[\neg M_{t-1}] - \eta(0 + \lambda\theta_t[\neg M_{t-1}]) \\ 
        &= (1-\eta\lambda)\theta_t[\neg M_{t-1}].
    \end{aligned}
\end{equation}
This suggests that if the model parameters $\theta$ is partitioned among all workers, we can avoid the expensive \textit{all-gather} operation for $\theta_{t+1}[\neg M_{t-1}]$ by computing it from each worker's local FP32 copy of the complete model (see \autoref{fig:dist-optm}). To reduce the memory overhead of keeping full model parameters on every worker, we offload the full-parameter replica to CPU memory. During parameter updates, we use a double-buffered pipeline that overlaps asynchronous CPU-to-GPU prefetch and sparse payload communication with local non-topk weight-decay updates, then writes back the updated buffer and offloads it to the host memory for the next iteration. 

\begin{algorithm}
    \caption{Optimize Megatron-LM distributed optimizer with SCAPE}
    \label{algo:mega-zero-opt}
    \begin{algorithmic}[1]
        \Require model parameter full FP32 copy $w$, parameter for forward and backward computation $\hat{w}$, parameter for communication $\tilde{w}$, local parameter $w^\prime$ updated by partitioned optimizer states, top-$k$ mask $M$, learning rate $\eta$, weight decay $\lambda$, 
        worker rank $n$, total number of workers $N$, step $t$, dataset $D$
        \Function{ForwardPass}{}
            \State $\tilde{w}_t^n \gets w^\prime_t[M_{t-1}]$ \Comment{Compress comm volume}
            \State {\color{red}$\hat{w}_t[M_{t-1}]\gets \text{AllGather}(\tilde{w}_t^n)$}
            \State {\color{red}$\hat{w}_t[\neg M_{t-1}]\gets (1-\eta\lambda)\hat{w}_t[\neg M_{t-1}]$ \Comment{Use FP32 $w_t$ to compute untransmitted params}}
            \State ForwardPass($D_t^n$, $\hat{w}_t$)
        \EndFunction
        \Function{BackwardPass}{}
            \State {\color{red}Start AsyncMaskRefresh($M_t$) \Comment{Hide mask sync with BW computation}} 
            \State $g_t^n\gets$BackwardPass($D_t^n$, $\hat{w}_t$) 
        \EndFunction
        \State {\redstrike{ReduceScatter($g_t^n$)}}
        \State $b_t^n\gets$ComputeUpdatedMoments($g_t^n$) 
        \State $\bar{b}_t\gets$AllReduce($b_t^n[M_{t-1}]$)
        \State UpdateLocalOptimizerStatesAndParams($\bar{b}_t$) 
    \end{algorithmic}
\end{algorithm}

\autoref{algo:mega-zero-opt} shows the optimized Megatron-LM with a distributed optimizer. 
If the model parameters are partitioned and need to be gathered across all workers before the forward- and backward-pass, we can reduce such communication volume by the inverse of the sparsity rate $1/k$. 
Additionally, to hide the communication cost of synchronizing and updating $M_{t}$, we launch the asynchronous \textit{all-gather} for $M_{t}$ after the backward-pass's \textit{all-gather} finishes, guaranteeing no communications interference. After the backward pass finishes, when workers synchronize their local temporal, semi-updated first-moment $b_t^n$, we replace the \textit{reduce-scatter} with the \textit{all-reduce} operation, because the LayerNorm, RMSNorm, and bias layers are usually not sparsified as they are less stable than the weights, leading to uneven buffers for \textit{reduce-scatter} on workers. 

\subsection{Optimization for Memory consumption}
SCAPE holds one additional residual buffer for error feedback, which has the same size as the model and uses FP32 for accuracy. It also needs two buffers for the masks of the current and next steps. Thus, the total memory overhead can be expressed as $\text{Mem}_{\text{overhead}} = \text{Mem}_{\text{cur\_mask}} + \text{Mem}_{\text{next\_mask}} + \text{Mem}_{\text{residual}}=W+W+4W=6W$~bytes,
where $W$ is the number of parameters, and the dtype for top-$k$ masks and residual are INT8 and FP32. 

To solve additional memory bottlenecks, we used CPU offloading with double buffering for residual and full parameters, minimizing the influence on both per-step time and memory. Furthermore, each value in a mask is represented as one bit, and we pack eight of them into one byte. Note that the smallest dtype supported by PyTorch operations is INT8, and this single-bit format is not supported. Therefore, we store both masks in packed single-bit format and use double buffers to convert when needed. We also use Triton kernels for packing and unpacking operations to reduce their computation overhead. After applying these memory optimization techniques, the memory overhead is $2\times W/8 + 2P + 8P=W/4+10P$ bytes, where $P$ represents the number of parameters in the largest layer, $2P$ represents the memory usage for double buffers for masks in INT8, and $8P$ represents the double buffers for the residual offloading. 

\section{Experiment Results}
In this section, we evaluate SCAPE from three complementary perspectives. We first present the pre-training, downstream, and end-to-end wall-clock results for Llama-500M, our primary modern LLM workload for demonstrating practical systems benefit. We then report GPT-345M results to show that SCAPE generalizes beyond a single architecture family and training configuration. Finally, we analyze per-step time and strong-scaling efficiency for Llama-500M and Llama-1.8B under Megatron-LM’s distributed optimizer, including the SCAPE-specific optimizations, from 4 to 64 GPUs.

\subsection{Experiment Setup}
We evaluate SCAPE by pre-training GPT-345M and Llama-500M on 32 NVIDIA GH200 GPUs of the Vista supercomputer~\cite{ruhela2024grace} at the Texas Advanced Computing Center (TACC). Each Vista node consists of a Grace-Hopper architecture with one GH200 GPU, 96~GB of HBM3 memory, and an NVLink-C2C interconnect between the Grace CPU and Hopper GPU. The nodes are connected through a 400~Gbps NVIDIA NDR InfiniBand network. To ensure a consistent and reproducible software environment, all experiments were conducted using the NVIDIA NGC PyTorch container (v26.01).

We evaluate the pre-trained models using zero-shot downstream benchmarks using lm-evaluation-harness~\cite{eval-harness}, including ARC (Easy and Challenge)~\cite{allenai:arc}, LAMBADA~\cite{paperno2016lambadadatasetwordprediction}, HellaSwag~\cite{zellers2019hellaswagmachinereallyfinish}, MMLU~\cite{hendrycks2021measuringmassivemultitasklanguage}, PIQA~\cite{bisk2019piqareasoningphysicalcommonsense}, WinoGrande~\cite{ai2:winogrande}, OpenBookQA~\cite{OpenBookQA2018}, and SuperGLUE~\cite{wang2020supergluestickierbenchmarkgeneralpurpose}.

\subsection{Llama-500M Pre-training}
\subsubsection{Model Architecture and Hyperparameters} We use the same model architecture of H2O-Danube3-500M~\cite{pfeiffer2024h2odanube3technicalreport} to define our Llama-500M model and pre-trained it on \hyperlink{https://huggingface.co/datasets/DKYoon/SlimPajama-6B}{SlimPajama-6B}, a subdataset sampled from SlimPajama~\cite{cerebras2023slimpajama}.  We used the same tokenizer as Llama-7B~\cite{touvron2023llama2openfoundation} from \hyperlink{https://huggingface.co/meta-llama/Llama-2-7b/tree/main}{HuggingFace}. We pre-train Llama-500M for 100,000 steps and use a global batch size of 1024 and a sequence length of 4096 for each step. The hyperparameters for the optimizers are: peak learning rate $\eta=3\times10^{-4}$, the minimum learning rate $\eta_{\text{min}}=3\times10^{-5}$, cosine learning rate decay, learning rate warmup for 2,000 steps, $(\beta_1,\beta_2)=(0.9,0.95)$, RMSNorm $\epsilon=10^{-5}$, weight decay $\lambda=0.1$, and the gradient clip of 1. We use BF16 for parameters and FP32 for the gradient. For SCAPE, we use sparsity warmup to exponentially decrease density $d$ from 1 to 0.1 or 0.01, and we do not compress RMSNorm layers.



\begin{table}
  \centering
  \caption{Final training and validation loss and end-to-end (E2E) wall-clock time of pre-training Llama-500M}
  \label{tab:llama-final-loss}
  \begin{sc}
    \resizebox{0.49\textwidth}{!}{%
      \begin{tabular}{cccc}
        \toprule
        Method & \begin{tabular}[c]{@{}c@{}}E2E Time\\(Days)\end{tabular} & Train loss & Val. loss \\\midrule
        AdamW (dense \textit{all-reduce}) & 2.47 (1$\times$) & 2.10 & 2.30 \\
        AdamS (dense \textit{all-reduce}) & 2.47 (1$\times$) & 2.11 & 2.32 \\
        SCAPE ($d$ = 0.1) & 1.59 (1.55$\times$)& 2.13 & 2.30 \\
        SCAPE ($d$ = 0.01) & 1.40 (1.76$\times$) & 2.17 & 2.31 \\
        \bottomrule
      \end{tabular}
    }
  \end{sc}
\end{table}

\begin{figure}
  \centering
  \begin{subfigure}[b]{0.49\columnwidth}
    \centering
    \includegraphics[width=\linewidth]{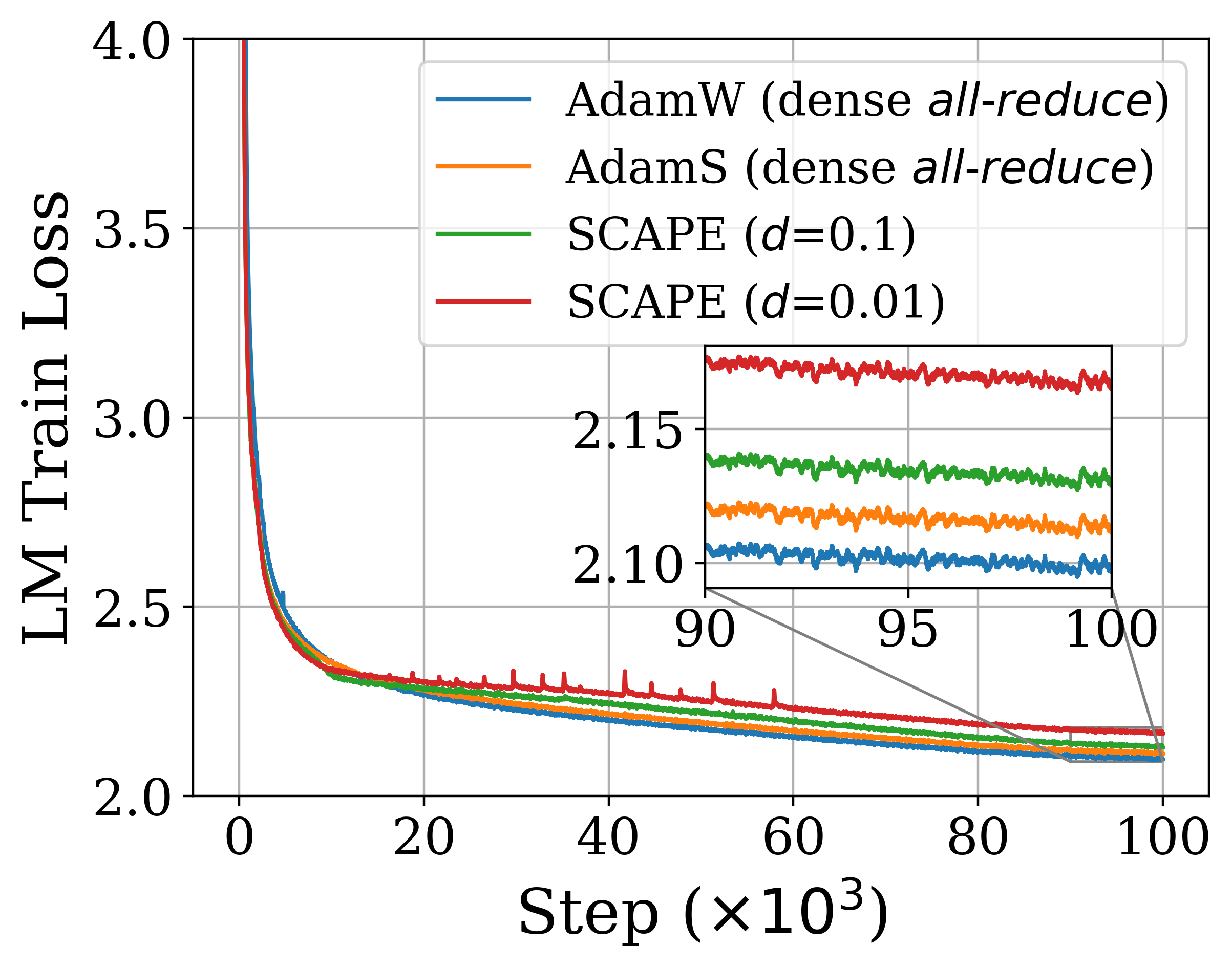}
    \caption{Training loss}
    \label{fig:llama-train-loss}
  \end{subfigure}
  \begin{subfigure}[b]{0.49\columnwidth}
    \centering
    \includegraphics[width=\linewidth]{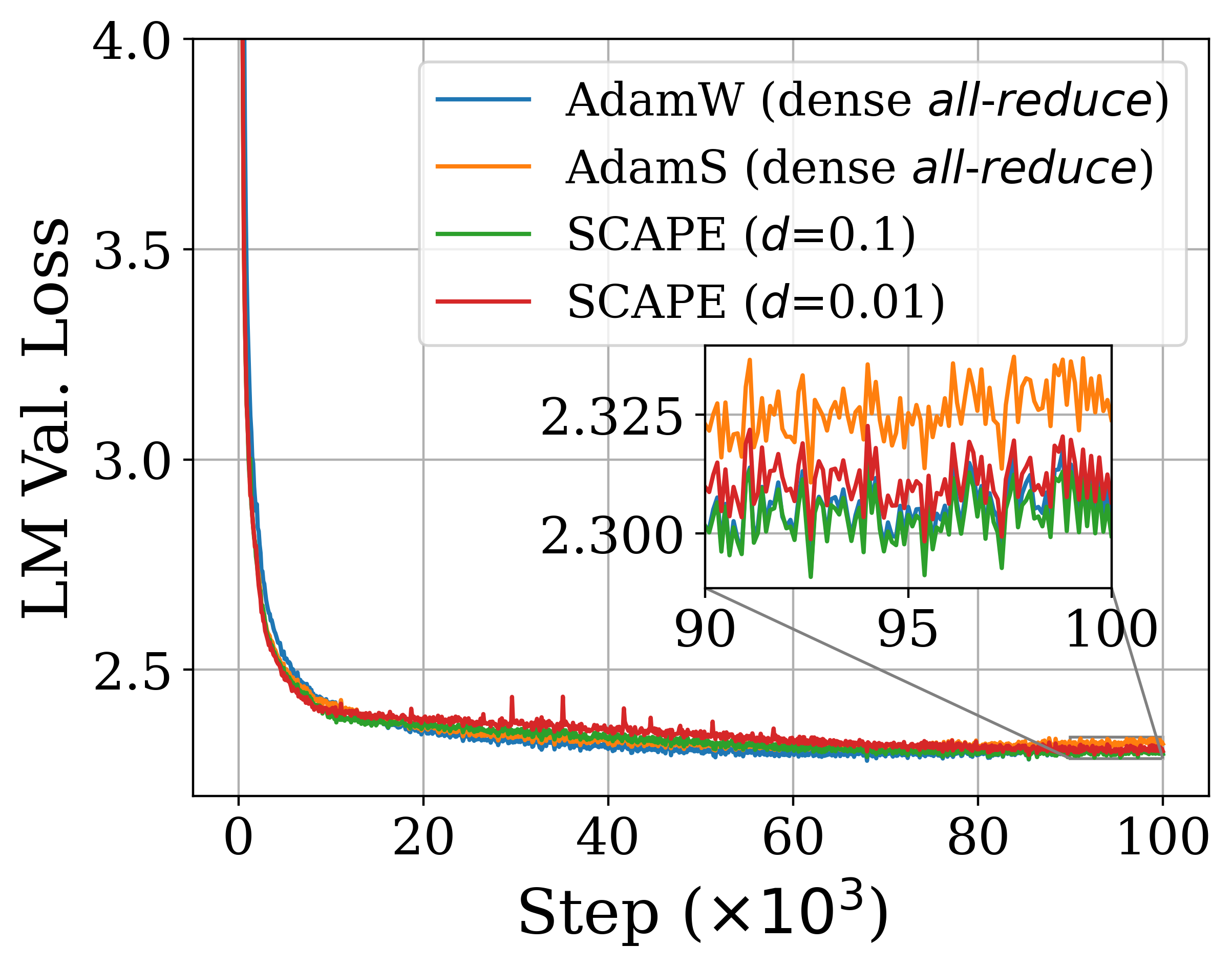}
    \caption{Validation loss}
    \label{fig:llama-val-loss}
  \end{subfigure}
  \caption{Pre-training loss curves for Llama-500M}
  \label{fig:llama-exp-res}
\end{figure}

\begin{table*}[t]
  \centering
  \caption{Zero-shot evaluation results for Llama-500M pre-trained using different approaches}
  \label{tab:llama-eval}
  \begin{sc}
          \begin{tabular}{ccccccccc}
        \toprule
        Method &
        \begin{tabular}[c]{@{}c@{}}ARC-C\\(ACC)\end{tabular} &
        \begin{tabular}[c]{@{}c@{}}ARC-E\\(ACC)\end{tabular} &
        \begin{tabular}[c]{@{}c@{}}HellaSwag\\(ACC)\end{tabular} &
        \begin{tabular}[c]{@{}c@{}}LAMBADA\\(ACC)\end{tabular} &
        \begin{tabular}[c]{@{}c@{}}MMLU\\(ACC)\end{tabular} &
        \begin{tabular}[c]{@{}c@{}}OpenBookQA\\(ACC)\end{tabular} &
        \begin{tabular}[c]{@{}c@{}}PIQA\\(ACC)\end{tabular} &
        \begin{tabular}[c]{@{}c@{}}WinoGrande\\(ACC)\end{tabular} \\
        \midrule
        AdamW (dense \textit{all-reduce})& \textbf{22.53} & 48.86 & \textbf{34.70} & \textbf{38.60} & 24.08 & 18.80 & \textbf{67.30} & 50.67 \\
        AdamS (dense \textit{all-reduce}) & 20.31 & 46.97 & 34.57 & 36.66 & \textbf{25.77} & 18.80 & 65.94 & \textbf{53.51} \\
        SCAPE ($d=0.1$) & 20.90 & \textbf{48.95} & 34.38 & 38.39 & 23.28 & \textbf{19.40} & 66.76 & 52.33 \\
        SCAPE ($d=0.01$) & 21.93 & 47.47 & 33.93 & 36.64 & 23.28 & 18.60 & 66.32 & 51.07 \\
        \bottomrule
        \end{tabular}%
  \end{sc}\vspace{5pt}
  \begin{sc}
      \begin{tabular}{cccccccccc}
        \toprule
        & \multicolumn{9}{c}{SuperGLUE} \\\cmidrule{2-10}
        Method &
        \begin{tabular}[c]{@{}c@{}}BoolQ\\(ACC)\end{tabular} &
        \begin{tabular}[c]{@{}c@{}}WiC\\(ACC)\end{tabular} &
        \begin{tabular}[c]{@{}c@{}}RTE\\(ACC)\end{tabular} &
        \begin{tabular}[c]{@{}c@{}}CB\\(ACC)\end{tabular} &
        \begin{tabular}[c]{@{}c@{}}ReCoRD\\(EM)\end{tabular} &
        \begin{tabular}[c]{@{}c@{}}WSC\\(ACC)\end{tabular} &
        \begin{tabular}[c]{@{}c@{}}MultiRC\\(ACC)\end{tabular} &
        \begin{tabular}[c]{@{}c@{}}COPA\\(ACC)\end{tabular} &
        Avg. \\
        \midrule
        AdamW (dense \textit{all-reduce}) & 57.65 & 49.69 & 53.43 & 35.71 & 69.81 & \textbf{41.35} & \textbf{57.03} & 65.00 & \textbf{53.71} \\
        AdamS (dense \textit{all-reduce}) & 57.55 & \textbf{50.63} & 53.07 & \textbf{41.07} & 69.54 & 34.62 & 49.86 & \textbf{68.00} & 53.04 \\
        SCAPE ($d=0.1$) & 51.90 & 49.06 & \textbf{57.40} & 28.57 & \textbf{70.00} & 39.42 & 53.11 & \textbf{68.00} & 52.18 \\
        SCAPE ($d=0.01$) & \textbf{57.92} & 50.00 & 52.71 & 32.14 & 69.66 & 36.54 & 56.48 & \textbf{68.00} & 52.93 \\
        \bottomrule
        \end{tabular}%
  \end{sc}
\end{table*}

\subsubsection{Pre-training Results}
The pre-training loss curves for Llama-500M using AdamW, AdamS, SCAPE ($d$ = 0.1), and SCAPE ($d$ = 0.01) are shown in \autoref{fig:llama-exp-res}. Their final end-to-end (E2E) wall-clock times, training losses, and validation losses are summarized in \autoref{tab:llama-final-loss}. Since AdamW is still the de facto optimizer for pre-training LLMs, we include it as a reference baseline. Compared with dense AdamS, SCAPE ($d$ = 0.1) reduces wall-clock time from 2.47 to 1.59 days (1.55$\times$ speedup, 35.6\% reduction) and lowers validation loss from 2.32 to 2.30. SCAPE ($d$ = 0.01) further reduces time to 1.40 days (1.76$\times$ speedup, 43.3\% reduction) with validation loss 2.31, which is still below AdamS. 
These results demonstrate SCAPE as a quality-preserving communication-efficient method for large-scale pre-training: it delivers substantial E2E wall-clock speedups while maintaining, and in validation loss slightly improving, model quality similar to dense AdamS.

\subsubsection{Downstream Task Evaluation}


\autoref{tab:llama-eval} shows that SCAPE largely preserves downstream task performance for the pre-trained Llama-500M despite using aggressive communication sparsity. Specifically, SCAPE ($d$ = 0.1) outperforms dense AdamW on 5 of 16 tasks and dense AdamS on 9 of 16 tasks.
In particular, SCAPE ($d$ = 0.1) surpasses both dense baselines on RTE and ReCoRD, while also improving over AdamW on ARC-E and OpenBookQA, and over AdamS on ARC-C, ARC-E, LAMBADA, OpenBookQA, PIQA, WSC, and MultiRC. Even at $d$ = 0.01, SCAPE remains competitive, exceeding AdamW on 4 of 16 tasks and AdamS on 7 of 16 tasks, including 3 SuperGLUE improvements over AdamW and 4 SuperGLUE improvements over AdamS. 
These results indicate that SCAPE retains strong downstream generalization for Llama-500M under 90\% and 99\% sparsity, with $d$ = 0.1 offering the best balance between compression and quality preservation.

\subsection{GPT-345M Pre-training}

\subsubsection{Model Architecture and Hyperparameters} We pre-train GPT-345M model using the same model architecture definition described in~\cite{radford2019language} on OpenWebText dataset~\cite{Gokaslan2019OpenWeb} for 100,000 steps. We set the global batch size to 512 and the sequence length to 1024. As for the hyperparameters of AdamW, AdamS, and SCAPE, we set $(\beta_1,\beta_2)=(0.9, 0.999)$, the peak learning rate $\eta=1.5\times10^{-4}$, the minimum learning rate $\eta_{\text{min}}=10^{-5}$, the learning rate scheduler to cosine, the learning rate warmup steps to 5,000 (5\% of the pre-training steps), the weight decay $\lambda=0.01$, and gradient clip to 1. Similar to Llama-500M training, mixed-precision training is used: the model parameters are in BF16, and the gradient is accumulated in FP32. Similarly, we use density warmup to gradually decrease $d$ from 1 to 0.1 or 0.01. The LayerNorm and bias layers in GPT-345M were not compressed.

\begin{figure}
  \centering
  \begin{subfigure}[b]{0.49\columnwidth}
    \centering
    \includegraphics[width=\linewidth]{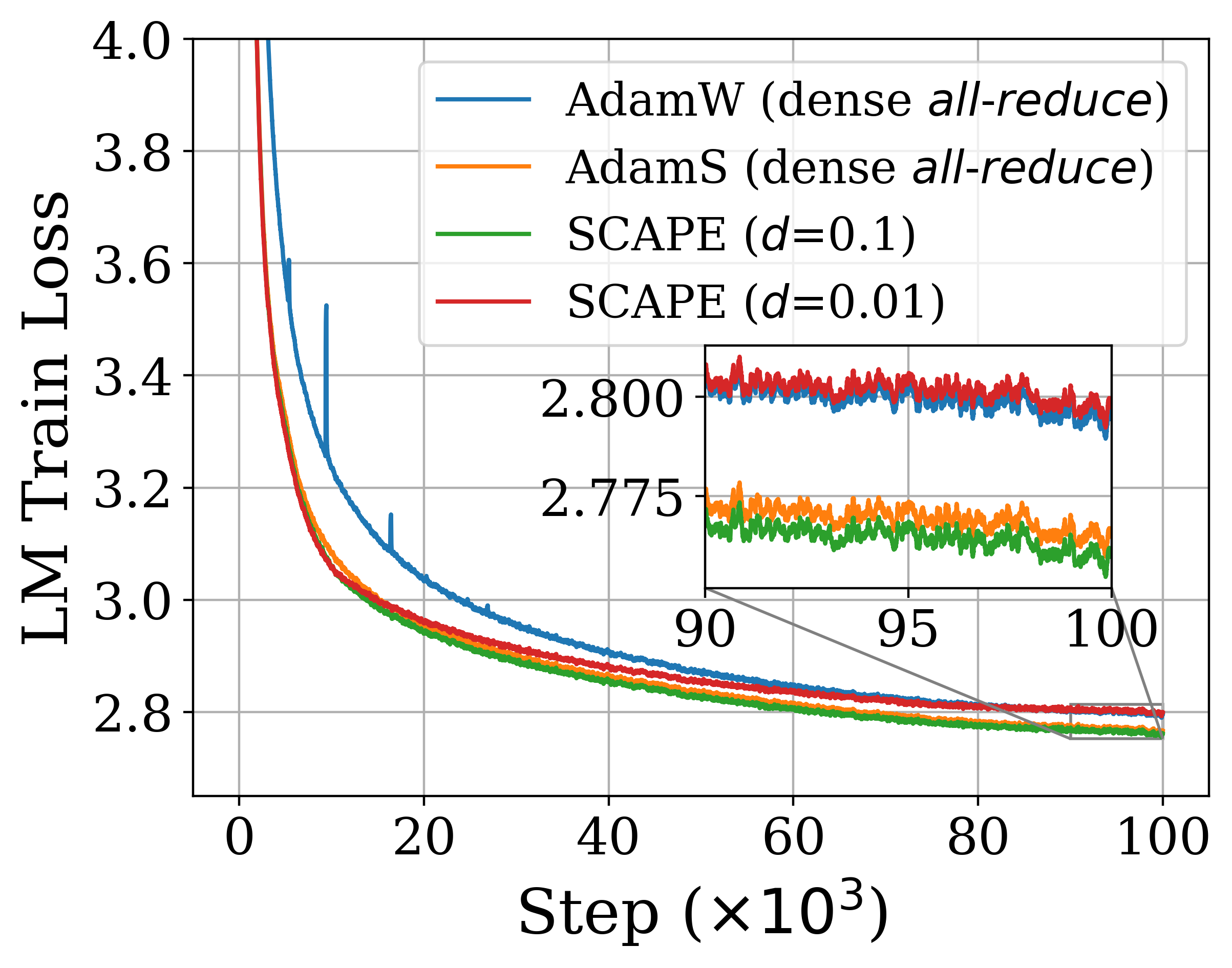}
    \caption{Training loss}
    \label{fig:gpt-train-loss}
  \end{subfigure}
  \begin{subfigure}[b]{0.49\columnwidth}
    \centering
    \includegraphics[width=\linewidth]{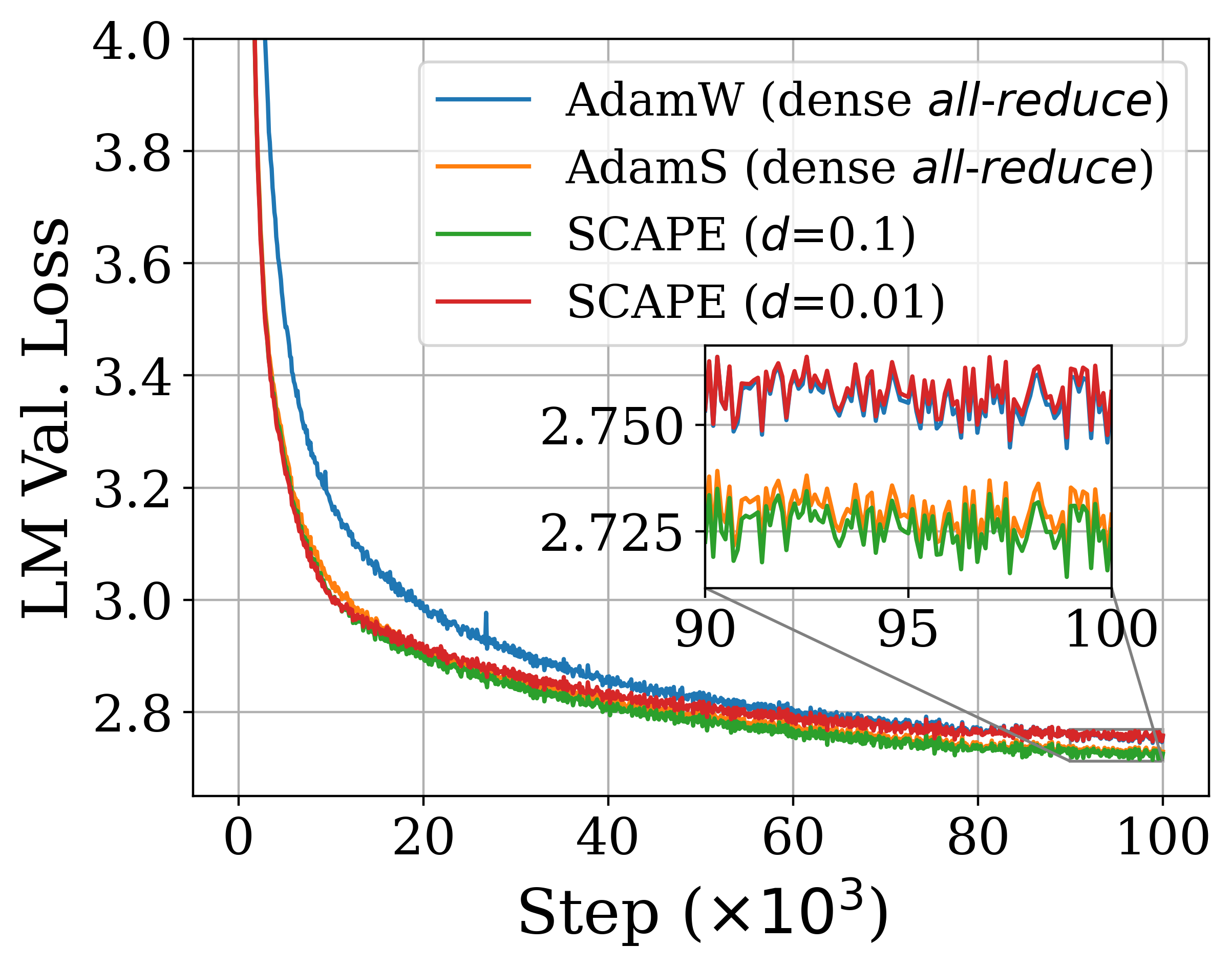}
    \caption{Validation loss}
    \label{fig:gpt-val-loss}
  \end{subfigure}
  \caption{Pre-training loss curves for GPT-345M}
  \label{fig:gpt-exp-res}
\end{figure}

\begin{table}
  \centering
  \caption{Final training and validation loss of pre-training GPT-345M}
  \label{tab:gpt-final-loss}
  \begin{sc}
    \begin{tabular}{ccc}
      \toprule
      Method & Train loss & Val. loss \\\midrule
      AdamW (dense \textit{all-reduce}) & 2.80 & 2.76 \\
      AdamS (dense \textit{all-reduce}) & 2.77 & 2.73 \\
      SCAPE ($d$ = 0.1) & 2.77 & 2.73 \\
      SCAPE ($d$ = 0.01) & 2.81 & 2.76 \\
      \bottomrule
    \end{tabular}
  \end{sc}
\end{table}

\begin{table*}[t]
  \centering
  \caption{Zero-shot evaluation results for GPT-345M pre-trained using different methods}
  \label{tab:gpt-eval}
  \begin{sc}
    \begin{tabular}{ccccccccc}
      \toprule
        Method &
        \begin{tabular}[c]{@{}c@{}}ARC-C\\(ACC)\end{tabular} &
        \begin{tabular}[c]{@{}c@{}}ARC-E\\(ACC)\end{tabular} &
        \begin{tabular}[c]{@{}c@{}}HellaSwag\\(ACC)\end{tabular} &
        \begin{tabular}[c]{@{}c@{}}LAMBADA\\(ACC)\end{tabular} &
        \begin{tabular}[c]{@{}c@{}}MMLU\\(ACC)\end{tabular} &
        \begin{tabular}[c]{@{}c@{}}OpenBookQA\\(ACC)\end{tabular} &
        \begin{tabular}[c]{@{}c@{}}PIQA\\(ACC)\end{tabular} &
        \begin{tabular}[c]{@{}c@{}}WinoGrande\\(ACC)\end{tabular} \\
        \midrule
        AdamW (dense \textit{all-reduce}) & 20.39 & \textbf{48.82} & 30.63 & 39.57 & 22.88 & 17.00 & 64.91 & 51.38 \\
        AdamS (dense \textit{all-reduce}) & 19.88 & 47.77 & \textbf{31.45} & 39.51 & \textbf{22.94} & 18.00 & \textbf{65.18} & \textbf{51.54} \\
        SCAPE ($d=0.1$) & \textbf{20.99} & \textbf{48.82} & 31.28 & \textbf{39.71} & 22.91 & 17.80 & 64.80 & 51.38 \\
        SCAPE ($d=0.01$) & 19.71 & 46.93 & 31.06 & 39.06 & 22.93 & \textbf{18.60} & 64.42 & 50.67 \\
        \bottomrule
    \end{tabular}\vspace{5pt}

    \begin{tabular}{cccccccccc}
    \toprule
    & \multicolumn{9}{c}{SuperGLUE} \\\cmidrule{2-10}
    Method & \begin{tabular}[c]{@{}c@{}}BoolQ\\(ACC)\end{tabular} & \begin{tabular}[c]{@{}c@{}}WiC\\(ACC)\end{tabular} & \begin{tabular}[c]{@{}c@{}}RTE\\(ACC)\end{tabular} & \begin{tabular}[c]{@{}c@{}}CB\\(ACC)\end{tabular} & \begin{tabular}[c]{@{}c@{}}ReCoRD\\(EM)\end{tabular} & \begin{tabular}[c]{@{}c@{}}WSC\\(ACC)\end{tabular} & \begin{tabular}[c]{@{}c@{}}MultiRC\\(ACC)\end{tabular} & \begin{tabular}[c]{@{}c@{}}COPA\\(ACC)\end{tabular} & \begin{tabular}[c]{@{}c@{}}Avg.\end{tabular} \\
    \midrule
    AdamW (dense \textit{all-reduce}) & 53.12 & 49.53 & 52.71 & 28.57 & 76.24 & \textbf{59.62} & \textbf{57.24} & \textbf{74.00} & \textbf{56.38} \\
    AdamS (dense \textit{all-reduce}) & 55.72 & 50.00 & \textbf{53.43} & \textbf{50.00} & \textbf{78.06} & 36.54 & 54.02 & 70.00 & 55.97 \\
    SCAPE ($d=0.1$) & \textbf{59.36} & \textbf{51.72} & 52.35 & 32.14 & 77.38 & 41.35 & 55.18 & 73.00 & 55.31 \\
    SCAPE ($d=0.01$) & 56.42 & 49.84 & 50.18 & 35.71 & 76.70 & 36.54 & 54.83 & 69.00 & 53.65 \\
    \bottomrule
    \end{tabular}%
  \end{sc}
\end{table*}

\subsubsection{Pre-training Results} The training and validation loss curves for pre-training GPT-345M model with AdamW, AdamS, SCAPE ($d$ = 0.1), and SCAPE ($d$ = 0.01) are shown in \autoref{fig:gpt-exp-res}, and the final training and validation loss are listed in \autoref{tab:gpt-final-loss}. 

Surprisingly, given the same token budget, when using $d=0.1$ (90\% sparsity), SCAPE achieves lower training and validation loss than the dense AdamS. When the sparsity rate is increased to 99\% (i.e., $d=0.01$), the final training and validation losses of SCAPE differ only slightly from those of AdamS, with differences below 0.04. Additionally, the difference between AdamW and SCAPE with $d=0.01$ is so small that it can be considered negligible.

\subsubsection{Downstream Task Evaluation}

\autoref{tab:gpt-eval} presents the zero-shot evaluation results of GPT-345M on the same downstream benchmark suite used for Llama-500M. Using dense AdamW as the reference baseline, SCAPE achieves strong task-level gains under aggressive sparsity. In particular, SCAPE ($d$ = 0.1) improves accuracy on 9 of 16 tasks, including ARC-C, HellaSwag, LAMBADA, MMLU, OpenBookQA, BoolQ, WiC, CB, and ReCoRD. Even when the sparsity is increased to $d$ = 0.01, SCAPE still outperforms AdamW on 8 of 16 tasks. These results indicate that GPT-345M is highly tolerant to SCAPE’s sparse synchronization, especially at $d$ = 0.1, where sparse training delivers improvements on a majority of the reported zero-shot evaluations while maintaining competitive overall downstream performance.

\begin{figure}
  \centering
  \begin{subfigure}[b]{0.9\columnwidth}
    \centering
    \includegraphics[width=\linewidth]{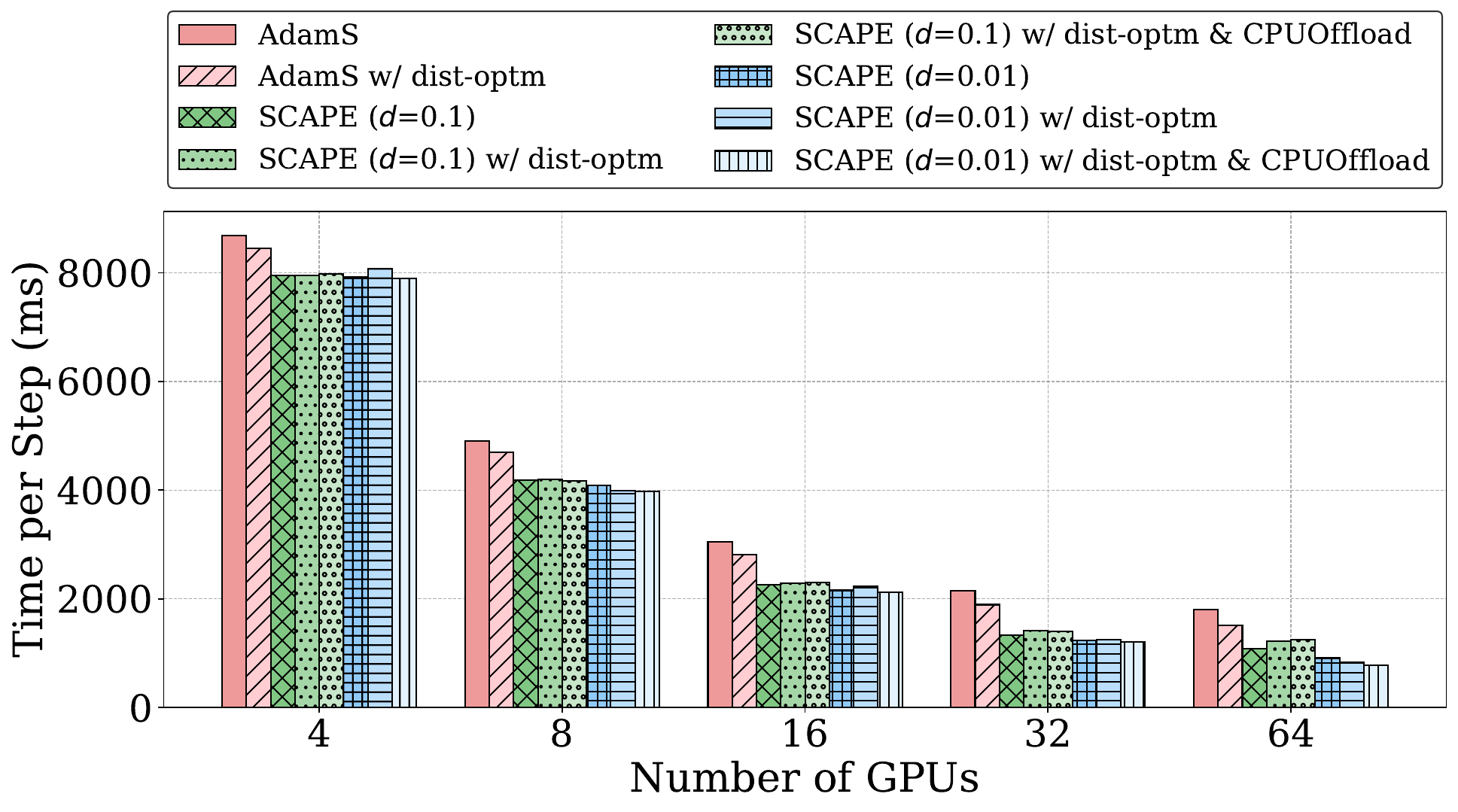}
    \caption{Llama-500M}
    \label{fig:llama-500M-per-step-time}
  \end{subfigure}
  \begin{subfigure}[b]{0.9\columnwidth}
    \centering
    \includegraphics[width=\linewidth]{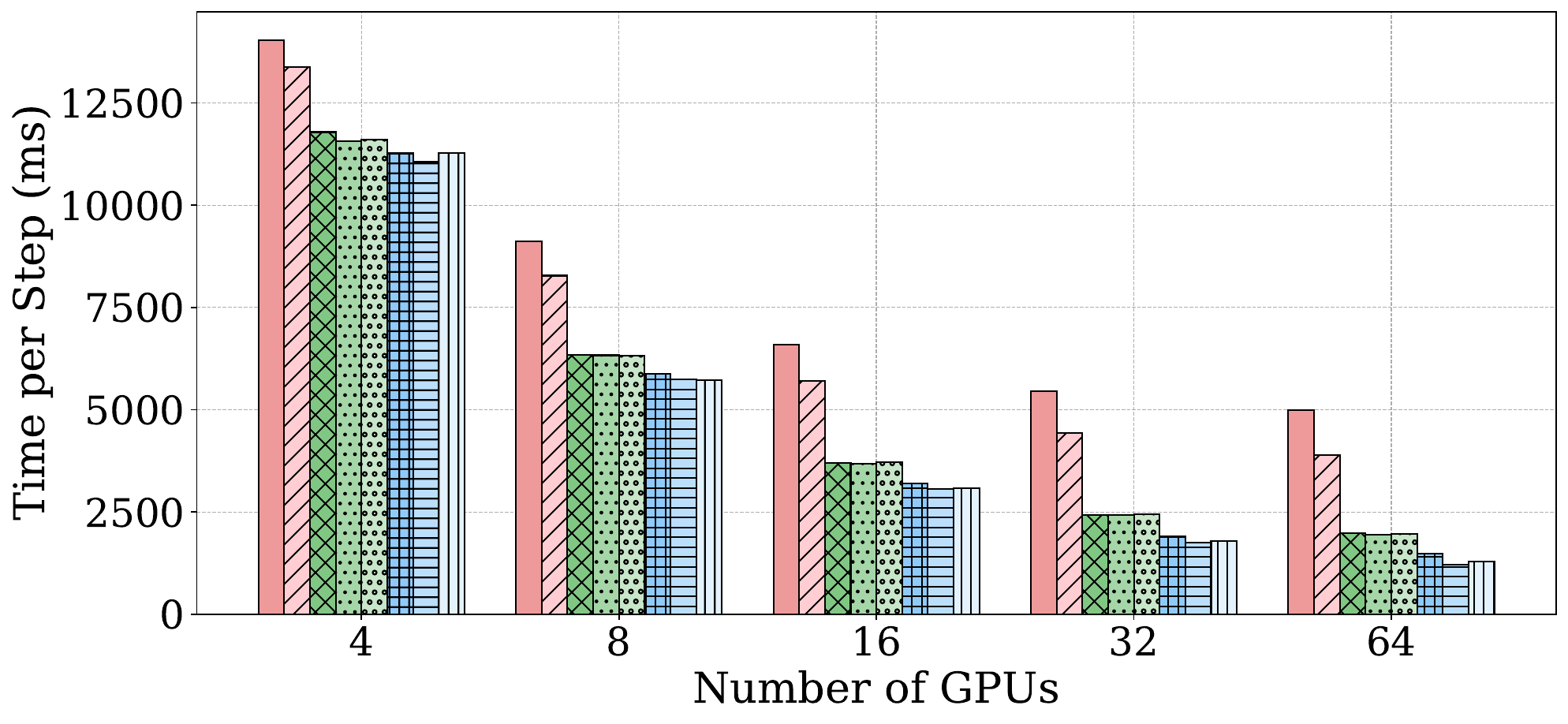}
    \caption{Llama-1.8B}
    \label{fig:llama-1.8B-per-step-time}
  \end{subfigure}
  \caption{Per-step time comparison between different methods used for training llama-500M (sequence length 4K) and Llama-1.8B (sequence length 2K)}
  \label{fig:per-step-time}
\end{figure}

\subsection{Per-step Time Analysis}
\label{sec:per-step-time}
We profile the time per iteration for pre-training Llama-500M (global batch 1024 and sequence length 4K) and Llama-1.8B (global batch size 1024 and sequence length 2K) with eight configurations.
These configurations include AdamS, AdamS with distributed optimizer, SCAPE ($d$ = 0.1), SCAPE ($d$ = 0.1) with distributed optimizer, SCAPE ($d$ = 0.1) with distributed optimizer and CPU offloading, SCAPE ($d$ = 0.01), SCAPE ($d$ = 0.01) with distributed optimizer, and SCAPE ($d$ = 0.01) with distributed optimizer and CPU offloading, where we define the model architecture of Llama-1.8B by following the definition of H2O-Danube3-1.8B~\cite{pfeiffer2024h2odanube3technicalreport}. The profiling results ranging from 4 to 64 GH200 GPUs are plotted in \autoref{fig:per-step-time}. 
Note that SCAPE ($d$ = 0.1) and SCAPE ($d$ = 0.01) refer to the DDP setting, and each worker has the full replicated model and optimizer states in GPU memory.  

As we can see from the profiling results, SCAPE can efficiently reduce the time per iteration for pre-training both Llama-500M and Llama-1.8B under different training configurations. For Llama-500M training, 
the per-step time reduction using SCAPE ($d$ = 0.1) and SCAPE ($d$ = 0.01) is less pronounced than Llama-1.8B. 
For Llama-500M on 64 GPUs, SCAPE with $d$ = 0.1 and $d$ = 0.01 reduces the per-step time from 1753.66~ms under dense AdamS to 1077.40~ms and 908.19~ms, corresponding to 1.63$\times$ and 1.93$\times$ speedup, respectively. For Llama-1.8B, SCAPE reduces the per-step time from dense AdamS's 4804.44~ms to 1983.08~ms and 1473.45~ms, yielding 2.42$\times$ and 3.26$\times$ speedup, respectively.
The reason is that with such a high number of tokens per step and a smaller model size than Llama-1.8B, the per-step training time for Llama-500M is dominated by computation rather than communication, which can be found from \autoref{fig:scaling-bottleneck}.


We notice that when the distributed optimizer is used, AdamS's per-step time is reduced, while SCAPE's is slightly increased for Llama-500M. The reason is that by sharding the model and optimizer states to all workers for AdamS, each worker's computation time for updating the model and states is reduced. Moreover, the composition of \textit{reduce-scatter} for synchronizing gradient and \textit{all-gather} for gathering the model shards in the distributed optimizer setting has the same cost as $T_{\text{ring-\textit{all-reduce}}}$ (see \autoref{eq:ring-allreduce}). Therefore, using distributed optimizer reduces the computation for each worker without introducing additional communication overhead. However, for SCAPE with distributed optimizer, since each worker still needs to update the full model, the computation is not reduced. Furthermore, the use of \textit{all-reduce}  introduces additional communication overhead compared to \textit{reduce-scatter} (see \autoref{eq:ring-allreduce}), although its communication volume is significantly reduced. Hence, using SCAPE with distributed optimizer has slightly higher per-step time than SCAPE when communication does not dominate the per-step time, such as training Llama-500M on four GPUs.

From \autoref{fig:per-step-time}, we can see that adding CPU offload to SCAPE with distributed optimizer does not add overhead for transmitting between CPU and GPU for each step. Thanks to the high bandwidth provided by GH200's NVLink-C2C of 900~GB/s, our double-buffering scheme for CPU offload only has minimal impact on the per-step time. Nevertheless, this could add extra overhead for systems that use PCIe for connecting CPU and GPU, which has a lower bandwidth than NVLink-C2C.

\begin{figure}
  \centering
  \begin{subfigure}[b]{0.9\columnwidth}
    \centering
    \includegraphics[width=\linewidth]{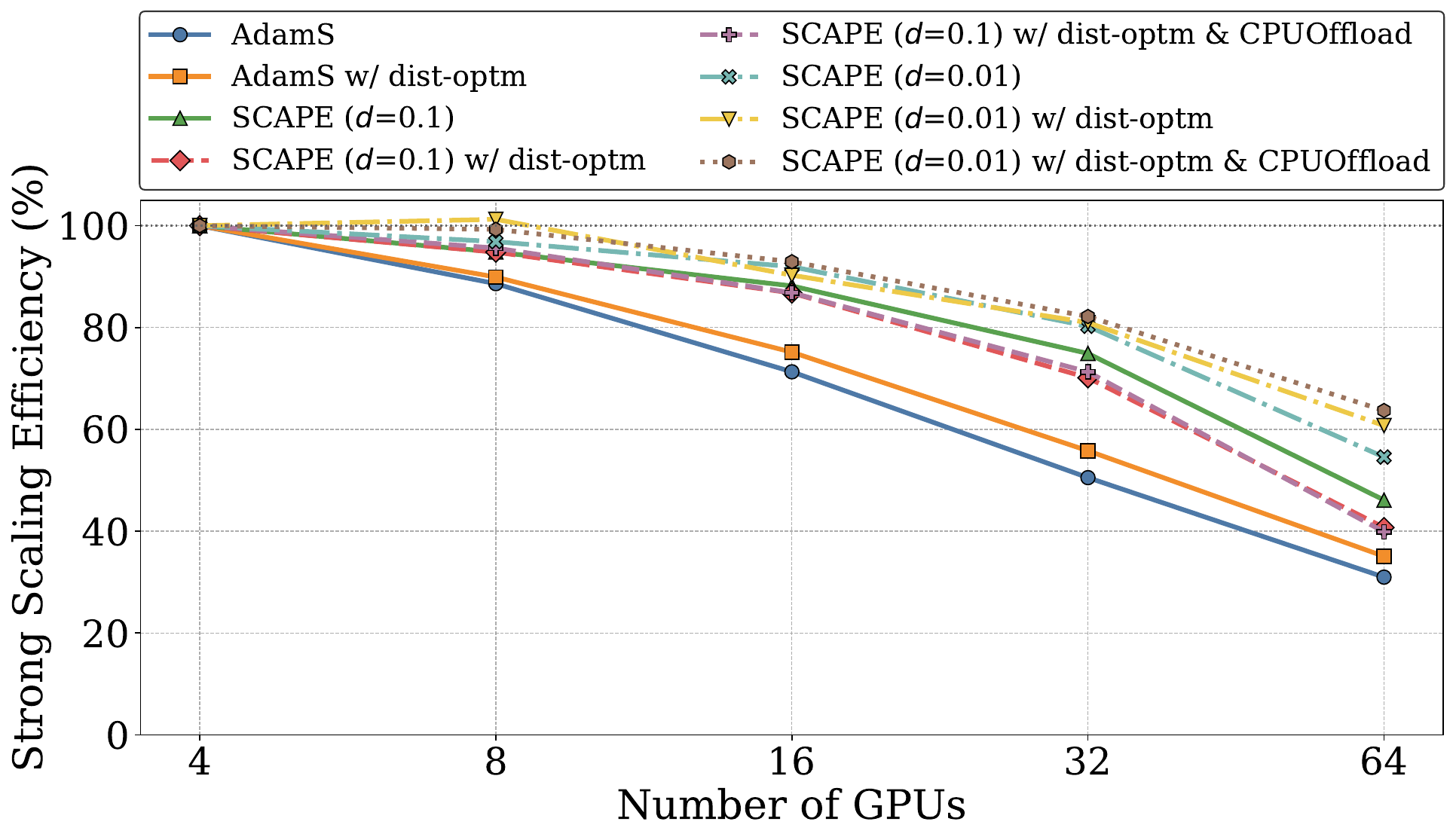}
    \caption{Llama-500M}
    \label{fig:llama-500M-strong-scaling}
  \end{subfigure}
  \begin{subfigure}[b]{0.9\columnwidth}
    \centering
    \includegraphics[width=\linewidth]{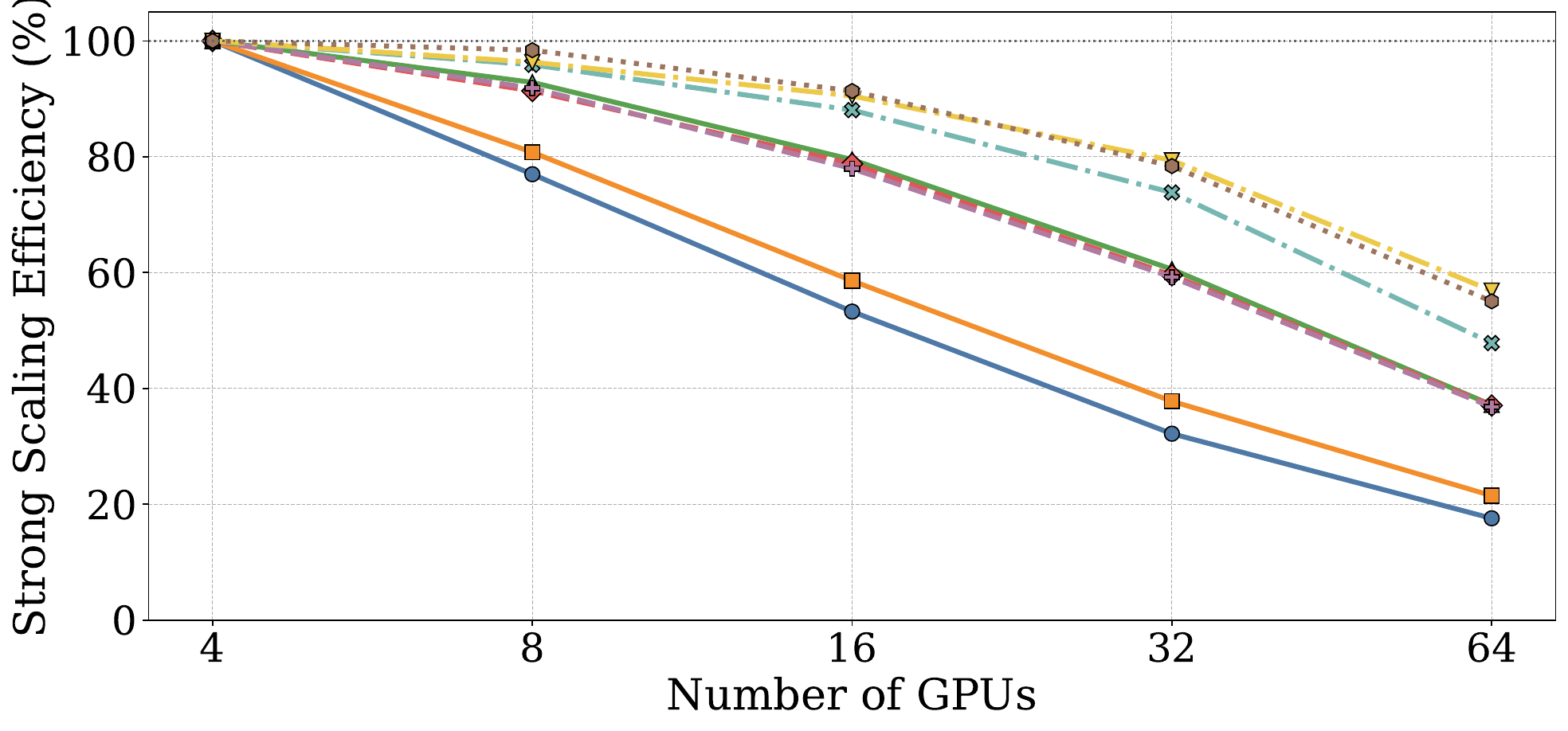}
    \caption{Llama-1.8B}
    \label{fig:llama-1.8B-strong-scaling}
  \end{subfigure}
  \caption{Strong scaling efficiency for training Llama-500M (sequence length 4K) and Llama-1.8B (sequence length 2K) with different methods}
  \label{fig:strong-scaling}
\end{figure}

\subsection{Strong Scaling Efficiency}
The strong scaling efficiency results for training Llama-500M and Llama-1.8B using all aforementioned methods on 4, 8, 16, 32, and 64 GH200 GPUs are plotted in \autoref{fig:strong-scaling}. 
Using the four-GPU setup as the baseline, the strong scaling efficiency improves with SCAPE for both models: for Llama-500M at 64 GPUs, efficiency is improved from AdamS's 30.98\% to 46.11\% for SCAPE ($d$ = 0.1) and 54.56\% for SCAPE ($d$ = 0.01), and up to 63.69\% when using SCAPE ($d$ = 0.01) with distributed optimizer and CPU offloading; 
for Llama-1.8B on 64 GPUs which has communication dominating the per-step time (see \autoref{fig:scaling-bottleneck}), the same pattern is more obvious, improving from 17.58\% for AdamS and 21.48\% for AdamS with distributed optimizer to 37.15\% for SCAPE ($d$ = 0.1), 47.83\% for SCAPE ($d$ = 0.01), and 55.04\% for SCAPE ($d$ = 0.01) with distributed optimizer and CPU offloading.

\begin{figure}
  \centering
  \begin{subfigure}[b]{0.49\columnwidth}
    \centering
    \includegraphics[width=\linewidth]{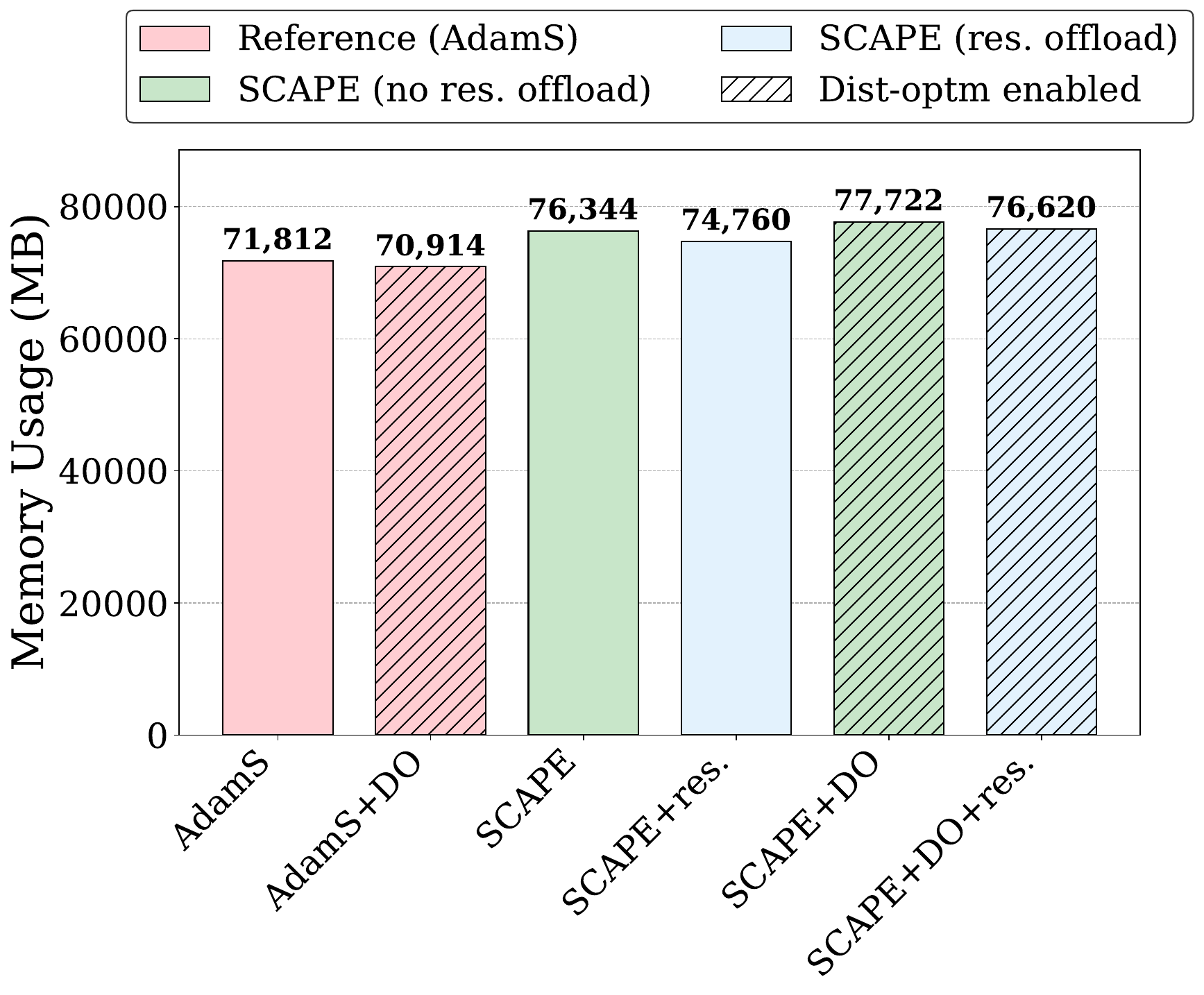}
    \caption{Llama-500M}
    \label{fig:llama-500M-memory}
  \end{subfigure}
  \begin{subfigure}[b]{0.49\columnwidth}
    \centering
    \includegraphics[width=\linewidth]{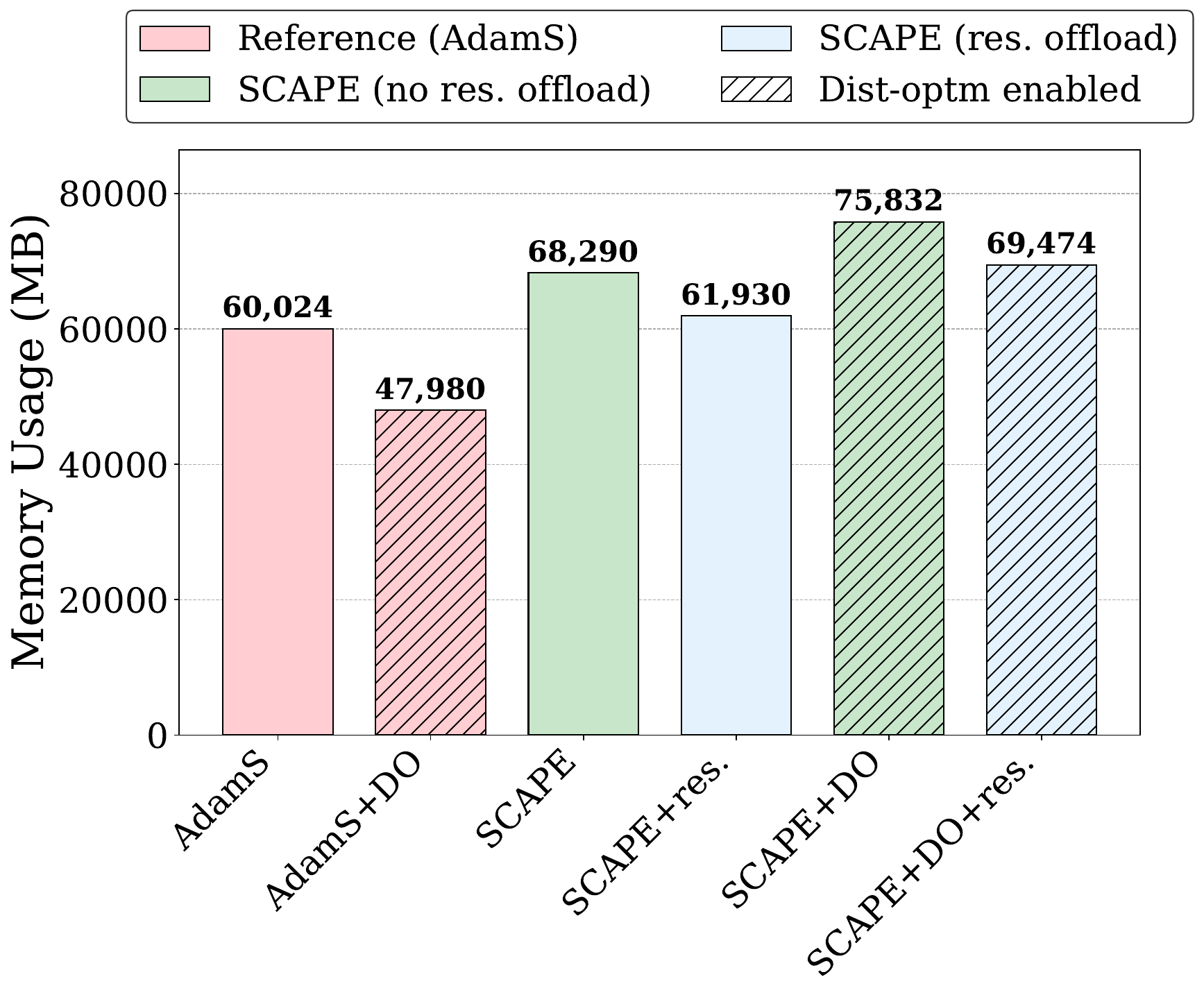}
    \caption{Llama-1.8B}
    \label{fig:llama-1.8B-memory}
  \end{subfigure}
  \caption{Memory usage for training Llama-500M (sequence length 4K) and Llama-1.8B (sequence length 2K)}
  \label{fig:memory}
  \vspace{-2ex}
\end{figure}

\subsection{Memory Consumption}
We profile the memory consumption for training Llama-500M and Llama-1.8B under different settings, and the results are shown in \autoref{fig:memory}. Since the memory overhead of residual buffer and full model parameters is determined by the number of model parameters and does not depend on SCAPE's $d$ or the number of workers, only the memory usage for $d$ = 0.1 on 4 GPUs is reported. We can see that with residual offloading enabled, the memory usage for training Llama-1.8B is reduced from 68,290~MB to 61,930~MB and from 75,832~MB to 69,474~MB before and after distributed optimizer is enabled. For Llama-500M training, because its model size is small and most of GPU memory is consumed by long-sequence activations, using residual offloading does not significantly reduce its memory usage.

\section{Related Work}
Many research works have been proposed to achieve communication-efficient pre-training through compression on gradient and model weights. Specifically, they can be classified into three main categories: gradient sparsification with error feedback, low-rank approximation, and quantization.

For gradient sparsification, DGC~\cite{lin2020deepgradientcompressionreducing} proposes a momentum-based sparsification method to reduce the communication volume of synchronizing gradient in DDP. DeMo~\cite{peng2026demodecoupledmomentumoptimization} proposes a decoupled momentum optimization method which first performs DCT to orthonormalize the momentum and then uses top-$k$ sparsification with error feedback to reduce communication volume. O$k$-Top$k$~\cite{li2022oktopk} proposes a novel sparse \textit{all-reduce} algorithm to achieve near asymptotically optimal communication volume. EDGC~\cite{yi2025edgcentropydrivendynamicgradient} proposes an entropy-driven adaptive gradient sparsification framework to dynamically specify the sparsity rate for each model layer. Radius~\cite{Radius} exploits the temporal stability of the top-$k$ gradient values' indices to amortize the computation cost of top-$k$ operations, and thus achieves lower computation overhead in gradient sparsification and higher throughput. 

For low-rank compression, ATOMO~\cite{wang2018atomocommunicationefficientlearningatomic} proposes to use singular value decomposition (SVD) to express gradients as atomic components and transmits only a subset of them to reduce communication overhead. PowerSGD~\cite{vogels2020powersgdpracticallowrankgradient} proposes a low-rank approximation of the gradient by representing it with two much smaller factor matrices. Optimus-CC~\cite{song2023optimusccefficientlargenlp} proposes a novel framework, combining 3D parallel training with PowerSGD.

For quantization-based methods, QSDP~\cite{markov2023quantizeddistributedtraininglarge} proposes to extend FSDP~\cite{zhao2023pytorchfsdpexperiencesscaling} with both model weight and gradient quantization to reduce the communication cost with convergence guaranteed. ZeRO++~\cite{wang2023zeroextremelyefficientcollective} improves the communication efficiency of ZeRO~\cite{rajbhandari2020zeromemoryoptimizationstraining} through the combination of block-wise quantized \textit{all-gather}, communication-aware data remapping, and quantized gradient averaging built on all-to-all communication 
SDP4Bit~\cite{jia2024sdp4bit4bitcommunicationquantization} proposes to use 4-bit quantization to reduce both weight and gradient communication via quantizing the weight differences and a two-level smooth quantization scheme for gradients, while also introducing runtime optimizations to mitigate quantization overhead.

\section{Conclusion and Future Works}
We present SCAPE, a communication-efficient distributed optimizer for LLM pre-training implemented based on Megatron-LM that achieves aggressive sparsification while preserving model quality on downstream evaluation tasks. 
Instead of sparsifying and transmitting the raw gradients, SCAPE leverages the stability of AdamS's first-moment to construct top-$k$ masks for sparse communication, aligns the top-$k$ mask generation with optimizer sharding, and applies the top-$k$ masks with a one-step delay to overlap the mask synchronization with computation. 
SCAPE also reconstructs the quantities required for second-moment updates from a single synchronized sparse buffer, thereby avoiding an additional collective. 
Through extensive evaluations, we show that SCAPE preserves the training loss, the validation loss, and the scores of the downstream task under both 90\% and 99\% sparsity. 
For Llama-500M, SCAPE reduces end-to-end pre-training wall-clock time by up to 43.3\% relative to dense baselines while maintaining comparable model quality. 
For Llama-1.8B, SCAPE achieves 3.26$\times$ speedup on 64 GPUs compared to dense AdamS.
In future work, we will explore extending SCAPE's moment-based sparsification to other optimizers, including Muon, which is gaining increasing popularity in production LLM training. 

\balance
\bibliographystyle{IEEEtran}
\bibliography{references}

\end{document}